\documentclass[10pt,twocolumn,letterpaper]{article}

\usepackage{cvpr}              
\usepackage[table]{xcolor}  
\usepackage{adjustbox}      
\usepackage{makecell}  
\usepackage{tabularx}
\usepackage[table]{xcolor}  
\usepackage{pifont}         
\usepackage{booktabs}
\newcommand{\cmark}{\ding{51}} 

\definecolor{cvprblue}{rgb}{0.21,0.49,0.74}
\usepackage[pagebackref,breaklinks,colorlinks,allcolors=cvprblue]{hyperref}

\title{Seeing Beyond: Extrapolative Domain Adaptive Panoramic Segmentation}

\author{Yuanfan Zheng$^{1}$ \qquad Kunyu Peng$^{2,3}$ \qquad Xu Zheng$^{4}$ \qquad Kailun Yang$^{1,}$\thanks{Corresponding author (e-mail: {\tt kailun.yang@hnu.edu.cn}).}
\\
\small $^{1}$Hunan University \quad $^{2}$Karlsruhe Institute of Technology \quad $^{3}$INSAIT, Sofia University ``St. Kliment Ohridski'' \quad $^{4}$HKUST(GZ)
}

\begin{document}
\maketitle
\begin{abstract}
Cross-domain panoramic semantic segmentation has attracted growing interest as it enables comprehensive $360^{\circ}$ scene understanding for real-world applications. However, it remains particularly challenging due to severe geometric  Field of View (FoV) distortions and inconsistent open-set semantics across domains. In this work, we formulate an open-set domain adaptation setting,  and propose \textbf{E}xtrapolative \textbf{D}omain \textbf{A}daptive \textbf{P}anoramic \textbf{S}egmentation (\textbf{E}\textbf{D}\textbf{A}-\textbf{P}\textbf{S}eg) framework that trains on local perspective views and tests on full $360^{\circ}$ panoramic images, explicitly tackling both geometric FoV shifts across domains and semantic uncertainty arising from previously unseen classes. To this end, we propose the Euler-Margin Attention (EMA), which introduces an angular margin to enhance viewpoint-invariant semantic representation, while performing amplitude and phase modulation to improve generalization toward unseen classes. Additionally, we design the Graph Matching Adapter (GMA), which builds high-order graph relations to align shared semantics across FoV shifts while effectively separating novel categories through structural adaptation. Extensive experiments on four benchmark datasets under camera-shift, weather-condition, and open-set scenarios demonstrate that \textbf{E}\textbf{D}\textbf{A}-\textbf{P}\textbf{S}eg achieves state-of-the-art performance, robust generalization to diverse viewing geometries, and resilience under varying environmental conditions. The code is available at \url{https://github.com/zyfone/EDA-PSeg}.
\end{abstract}    
\begin{figure}
    \centering
    \includegraphics[width=1.0\linewidth]{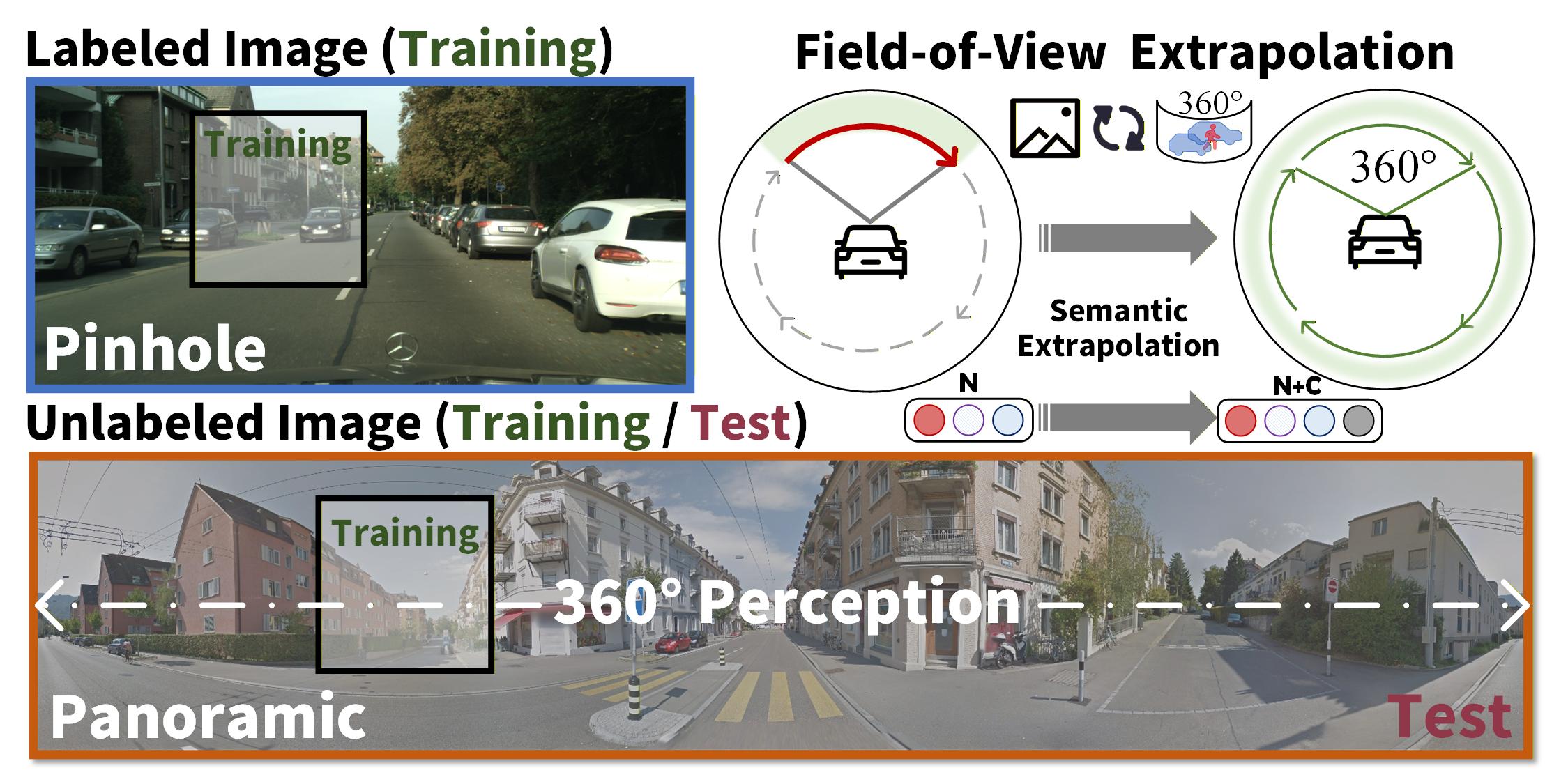}
\caption{\textbf{Extrapolative Domain Adaptation (EDA)} extends beyond the training FoV and known semantic categories, facilitating the transfer of knowledge from \textit{pinhole supervision}  to {unlabeled 360\textdegree} perception. It addresses both cross-view \textbf{geometric distortions} and the \textbf{semantic extrapolation} to unknown categories.
}
\label{task}
\end{figure}

\section{Introduction}
\label{sec:intro}
Panoramic vision~\cite{lin2025one,ai2025survey_omnidirectional} provides an omnidirectional perspective with a 360\textdegree{} Field of View (FoV), enabling occlusion-aware and seamless scene perception~\cite{cao2024occlusion,cao2025unlocking,liu2025prior}.
Cross-domain Panoramic Segmentation (CPS)~\cite{ma2021densepass,zhang2022bending,zheng2023both} addresses the domain shift induced by FoV differences between conventional pinhole and panoramic images. 
In the Unsupervised Domain Adaptation (UDA) setting~\cite{wilson2020survey}, CPS models are trained on labeled pinhole images (source domain) and adapted to unlabeled 360\textdegree{} panoramic images (target domain). 
By alleviating annotation costs and overcoming the limited FoV of pinhole cameras, CPS facilitates robust wide-angle semantic scene understanding for applications in autonomous driving~\cite{ma2021densepass,yang2021capturing} and robotics~\cite {wu2025quadreamer,bacchin2024pannote}.
\par
Most existing CPS methods~\cite{zhang2024goodsam,zheng2023both,zhong2025omnisam} operate under a closed-set assumption, where all test instances belong to the same classes observed during training. Under this setting, these methods generally adopt two complementary strategies to handle domain shifts across camera views. The first strategy focuses on FoV alignment~\cite{zheng2023look,zheng2023both}, aiming to reduce distortions introduced by variations in camera FoV. Representative approaches address this issue through geometric projection adaptation~\cite{zheng2023look} or sliding-window patch processing~\cite{zhang2024goodsam,zheng2023both,zhong2025omnisam}. The second strategy focuses on semantic alignment between the source and target views~\cite{zhang2024behind,zhong2025omnisam}, adapting the category prototype features to match the category distribution in closed-set scenarios.
\par
Current CPS methods perform well in controlled environments but often struggle in open-world scenarios~\cite{choe2024open,choe2025universal}. This limitation is particularly critical in autonomous driving, where vehicles frequently encounter unseen objects outside their field of view, posing significant safety risks. While Open Set Domain Adaptation (OSDA)~\cite{geng2020recent, fang2020open} offers potential solutions, existing approaches such as BUS~\cite{choe2024open} and UniMAP~\cite{choe2025universal} rely on constructing category prototypes for pixel-level mapping or prototype weight scaling. Panoramic images suffer from style inconsistencies and geometric distortions~\cite{zheng2023look}, making them ill-suited for existing pinhole-domain Open-Set UDA methods designed for pixel-level domain alignment. The challenge is further amplified under diverse weather conditions~\cite{liu2020open,yao2025scmix,zhao2022source}, where adapting from one or multiple weather conditions to diverse adverse weather scenarios disrupts cross-domain alignment.
\par
In this paper, we formulate an open-set cross-domain panoramic semantic segmentation, which aims to enable models to generalize to unseen categories while adapting to diverse FoV scenes under varying weather conditions. To the best of our knowledge, this is the first study to address this challenging yet practical problem, which is crucial for achieving comprehensive and reliable scene understanding in unconstrained real-world environments. To this end, we propose \textbf{E}xtrapolative \textbf{D}omain \textbf{A}daptive \textbf{P}anoramic \textbf{S}egmentation (\textbf{E}\textbf{D}\textbf{A}-\textbf{P}\textbf{S}eg) as shown in \cref{task}. This novel framework emphasizes extrapolating knowledge learned from conventional pinhole-view images to omnidirectional 360{\textdegree} panoramic scenes, thereby facilitating both cross-view generalization and open-set domain adaptation.
\par
Specifically, \textbf{E}\textbf{D}\textbf{A}-\textbf{P}\textbf{S}eg consists of two core components, the Graph Matching Adapter (GMA) and Euler-Margin Attention (EMA). GMA samples and synthesizes graph nodes to model high-order class relations, overcoming the limitations of pixel-level prototype representations and performing open-set graph matching with regularization to separate known and unknown classes. Euler-Margin Attention (EMA) employs an Euler formula-based transformation to project features into an angle-aware embedding space. It achieves amplitude and phase modulation through learnable parameters, enabling adaptive adjustment of the amplitude distribution and phase scaling according to the semantic angle for cross-view generalization. The framework is evaluated under diverse domain shifts, including camera and weather conditions. We conduct extensive experiments on four benchmarks, covering five datasets, which encompass \textit{Camera Shift} and \textit{Weather Shift}. Our method consistently outperforms existing approaches, achieving the improvement of 3.39\% mIoU in Cityscapes $\rightarrow$ DensePASS benchmark. Furthermore, we systematically evaluate and benchmark existing CPS methods alongside OSDA methods under the \textbf{E}\textbf{D}\textbf{A}-\textbf{P}\textbf{S}eg setting. These results not only verify the effectiveness of our approach in cross-view generalization but also demonstrate its superior generalization across open-set and adverse weather conditions. Our main contributions are summarized as follows:
\begin{itemize}
\item 
This paper formulates a practical cross-view setting for Open-Set UDA panoramic semantic segmentation and proposes \textbf{Extrapolative Domain Adaptive Panoramic Segmentation (\textbf{E}\textbf{D}\textbf{A}-\textbf{P}\textbf{S}eg)}, a framework that simultaneously mitigates FoV-induced geometric distortions and enables semantic extrapolation to unseen viewpoints.
\item 
We propose a Graph Matching Adapter (GMA) module that models high-order semantics for aligning known classes while effectively distinguishing novel categories.
\item 
We further propose the Euler-Margin Attention (EMA) module, which enhances semantic angle and improves cross-view generalization to FoV shifts and novel classes.
\item Extensive experiments on multiple benchmarks, including Cityscapes, ACDC, DensePASS, GTA, and SynPASS, demonstrate that ours achieves state-of-the-art performance under diverse domain shifts, establishing an effective baseline for Open-Set UDA panoramic segmentation.
\end{itemize}

\section{Related Work}
\label{sec:relate}

\noindent\textbf{Cross-Domain Panoramic Segmentation (CPS).}
Existing CPS methods~\cite{ma2021densepass,zhang2022bending,zheng2023both} adopt UDA settings to bridge the domain gap between synthetic or real pinhole images and 360\textdegree{} panoramic images. Most approaches focus on geometric field-of-view alignment and semantic prototype alignment. For distortion mitigation, CFA~\cite{zheng2023look} introduces distortion-aware attention to capture pixel distribution discrepancies, while DPPASS~\cite{zheng2023both} employs cross-domain and intra-projection training with tangent projections. Trans4PASS~\cite{zhang2024behind} further addresses geometric distortions through deformable patch embeddings and MLP-based adaptation. In terms of semantic alignment, OmniSAM~\cite{zhong2025omnisam} leverages SAM~\cite{kirillov2023segment} for prototype alignment, and GoodSAM~\cite{zhang2024goodsam} incorporates SAM boundary priors within a knowledge distillation framework. Additional studies~\cite{zheng2023both,zhong2025omnisam,zhang2024goodsam++} enhance semantic consistency via mutual prototypical adaptation and pseudo-label refinement. The broader frameworks~\cite{lin2025one,xu2025mamba4pass,yang2020omnisupervised,jaus2023panoramic} extend these ideas to foundation models and representation learning~\cite{zheng2024semantics,jiang2025multi,jiang2025gaussian,zhang2024behind}, with OPS~\cite{zheng2024open} providing further complementary advances. In the context of adaptation to the source-free domain, where source data is unavailable, recent methods~\cite{cao2024occlusion,cao2025unlocking,zhang2024goodsam,zhang2024goodsam++,zheng2024360sfuda++} emphasize self-training strategies and integration with SAM~\cite{kirillov2023segment}. In this paper, we propose EDA-PSeg, the first framework designed for Open-Set UDA CPS.

\begin{figure*}[!t]
    \centering
    \includegraphics[width=1.0\linewidth]{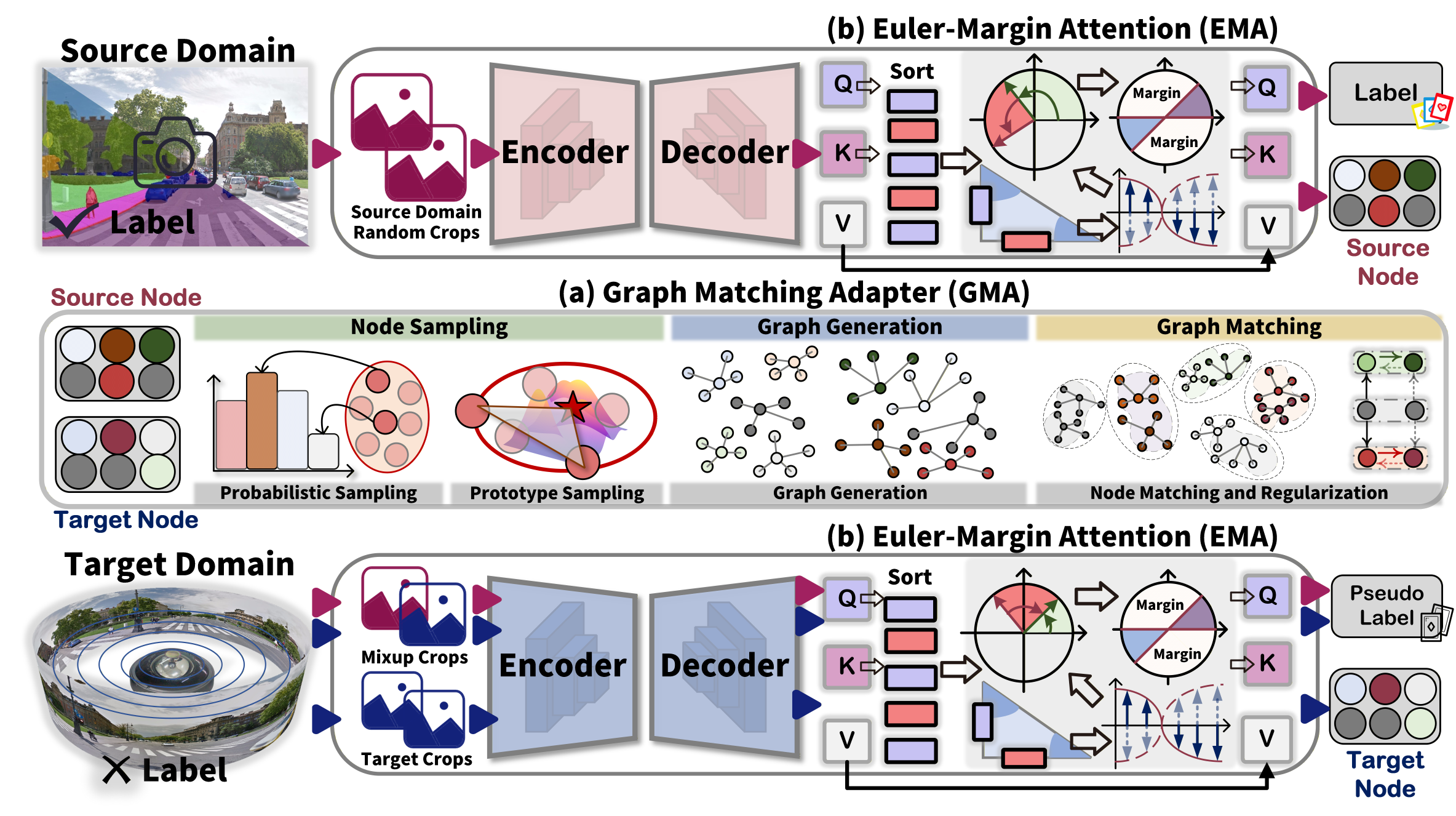}
    \caption{Illustration of the proposed \textbf{E}\textbf{D}\textbf{A}-\textbf{P}\textbf{S}eg, which tackles cross-view and semantic extrapolation. The framework incorporates two key components: the Graph Matching Adapter (GMA), which constructs a high-order graph to align domain shared class graph nodes, and the Euler-Margin Attention (EMA), which models features with the Euler formula to enhance angle invariance under unseen viewpoints.}
    \label{framework}    
\end{figure*}

\noindent\textbf{Positional Encoding for Self-Attention.} Positional encoding~\cite{chen2021simple} is a fundamental component of self-attention, allowing the model to capture semantic structures in both natural language processing~\cite{ke2020rethinking,zhao2023survey} and computer vision~\cite{cordonnier2019relationship,heo2024rotary}. Depending on how positional information is represented, existing schemes can be broadly categorized as absolute~\cite{vaswani2017attention} or relative~\cite{raffel2020exploring}. Absolute positional encoders assign unique position-dependent representations using sinusoidal or learned embeddings, while relative encoders model only positional offsets between query and key tokens to achieve translation invariance. CFA~\cite{zheng2023look}, for example, introduces a distortion-aware attention mechanism for the CPS task that leverages absolute positional encoding. Meanwhile, recent large language models, such as LLaMA, adopt Rotary Positional Embeddings (RoPE)~\cite{su2024roformer}, which integrate both absolute and relative positional information.
Building on this idea, EulerFormer~\cite{tian2024eulerformer} unifies semantic and positional representations in Euler space. In this work, we propose Euler-Margin Attention, a module for semantic extrapolation that robustly generalizes across cross-view geometric distortions via angle-margin projection and amplitude and phase modulation in complex vector space.

\noindent\textbf{Cross-domain Graph Matching.} Graph matching establishes topological correspondences among features by jointly optimizing node and edge affinities, thus enabling structured reasoning over relational dependencies. Recent advances in graph-based methods have substantially improved segmentation~\cite{chen2025bi,liu2023grab} through enhanced semantic alignment and boundary-aware feature interactions. 
Moreover, the cross-domain graph matching method has shown remarkable effectiveness across diverse tasks, including cross-domain named entity recognition~\cite{zheng2022cross}, medical image analysis~\cite{he2025gagm,lv2025test}, and object detection~\cite{chen2022relation,li2022scan++,li2023sigma++,liu2023cigar}.
Despite significant progress under the closed-set setting, existing cross-domain graph matching methods remain in their infancy in the open-set scenario.  In this paper, we effectively remodel the closed-set graph matching to support open-set graph matching. This enables handling of unknown-class nodes, bridging the gap for graph matching.

\section{Method}
\label{sec:method}

\subsection{Overview}
The architecture of the proposed \textbf{E}\textbf{D}\textbf{A}-\textbf{P}\textbf{S}eg is illustrated in \cref{framework}, highlighting its main components and processing pipeline. We consider a labeled source domain with pinhole images $\mathcal{D}_s = \{(\mathbf{x}_s^i, \mathbf{y}_s^i)\}_{i=1}^{\mathcal{B}}$, and an unlabeled target domain with panoramic images $\mathcal{D}_t = \{\mathbf{x}_t^i\}_{i=1}^{\mathcal{B}}$. Following the architectural overview, we define the label spaces for both domains. The source label space $\mathcal{Y}_s$ contains the known classes, while the target label space is defined as $\mathcal{Y}_t = \mathcal{Y}_s \cup \mathcal{Y}_u$, where $\mathcal{Y}_u$ denotes additional unknown classes. Random cropping is applied to source and mixed samples for label and pseudo-label supervision, while source-only and target-only crops are used for graph sampling. The cropped images are then fed into an encoder-decoder network, followed by the proposed Euler-Margin Attention (EMA). EMA constructs an Euler–Margin projection to embed features into a bounded angular space, mitigating cross-view geometric distortions while preserving the panoramic semantic structure. Moreover, EMA performs amplitude and phase modulation to enhance feature separability between known and unknown classes. The features are passed to the Graph Matching Adapter (GMA), which performs graph matching to geometrically decouple known classes while pushing unknown classes apart, thereby facilitating semantic extrapolation to unseen categories.

\subsection{Graph Matching Adapter (GMA)}
Given source $\mathcal{D}_s$ and target $\mathcal{D}_t$ domains, the encoder-decoder network extracts node features
$\mathbf{V}_{s,t} \in \mathbb{R}^{B \times N \times d}$, where $N = H \times W$ and $d$ is the feature channel dimension. The $s,t$ subscripts are omitted unless explicitly specified.
\\
\textbf{Node Sampling.} 
We perform local node sampling guided by confidence, entropy, and prototype distance to select nodes representing local semantics, and subsequently aggregate them into class-wise global prototypes to form robust global semantic representations.  Given the node features $\mathbf{V}$ from the encoder and decoder (flattened), we first apply the classifier head to obtain the predictions per-pixel. For each pixel, we compute a confidence score $\mathbf{p}_i$ and an uncertainty measure $\mathcal{H}(\mathbf{p}_i)$. For any base class index $b$, we define the positive set $\mathbf{S}_{\rm pos}^{(b)} = \{  \mathbf{V}_i \mid \mathbf{p}_i > \tau_p \wedge \mathcal{H}(\mathbf{p}_i) < \tau_e \}$ and the negative set $\mathbf{S}_{\rm neg}^{(b)} = \{  \mathbf{V}_i \mid \mathbf{p}_i \le \tau_p \wedge \mathcal{H}(\mathbf{p}_i) < \tau_e \}$, where $\tau_p$ and $\tau_e$ denote confidence and entropy thresholds, respectively, determined using percentile strategies with 0.5. We aggregate prototypes as $\mathbf{G}^{(b)} = \omega^{(b)} + \sigma$, where $\sigma$ is Gaussian noise, and $\omega^{(b)}$ is the mean of the same label feature. To refine node selection, we retain the $K$ nearest nodes to the prototype, defined as $\hat{\mathbf{S}}_{{\rm pos,neg}}^{(b)} = \{ \mathbf{V}_i \mid d_i \le d_{(K)}, \; d_i = \|\mathbf{V}_i - \omega^{(b)}\| \}$, where $d_i$ denotes the distance of the $i$-th node to the current prototype mean $\omega^{(b)}$. For 
the novel class index $n$, nodes are sampled based on similar confidence but distinct entropy values, specifically those satisfying $\mathcal{H}(\mathbf{p}_i) < \tau_m$ for the positive and $\mathcal{H}(\mathbf{p}_i) > \tau_m$ for negative samples, where $\tau_m$ is the median entropy of the distribution $\mathcal{H}(\mathbf{p})$. This procedure yields the sets $\hat{\mathbf{S}}_{{\rm pos,neg}}^{(n)}$ and their corresponding prototypes $\mathbf{G}^{(n)}$. Finally, by concatenating the base and novel class nodes along with their corresponding prototypes, we obtain the set $\mathbf{S}_\varepsilon$ as follows:
\begin{equation}
\small
\mathbf{S}_\varepsilon = \left\{
\hat{\mathbf{S}}_{{\rm pos,neg}}^{(b)},
\mathbf{G}^{(b)},
\hat{\mathbf{S}}_{{\rm pos,neg}}^{(n)},
\mathbf{G}^{(n)}
\right\},
\end{equation}
where $\mathbf{S}^{(b),(n)}_{\rm pos, neg}$ represents pixel-level nodes capturing class diversity, and $\mathbf{G}^{(b),(n)}$ denotes the node means representing the global class distribution in the batch data. We perform both local and global class sampling to facilitate graph matching for dense prediction, while avoiding excessive nodes that could lead to computational redundancy. We then update the global memory bank $\mathcal{M}$ using the candidate node samples $\mathbf{S}_\varepsilon$ via exponential moving average with parameter $\alpha$, where this process maintains a stable representation by weighting the new mean sample vector $\mathbf{V}_i$ and the existing memory, as
$\mathcal{M} \gets (1-\alpha)\, \mathcal{M} + \alpha \, \frac{1}{|\mathbf{S}_c|} \sum_{i \in \mathbf{S}_c} \mathbf{V}_i$,
applied independently for each class-specific set $\mathbf{S}_c$.
\\
\textbf{Graph Generation.} 
The Graph Generation module first identifies shared categories between the source and target domains, fills in missing categories using memory banks $\mathcal{M}$, and then builds node and graph affinities via self-attention~\cite{vaswani2017attention}. Given a candidate node set $\mathbf{S}_\varepsilon$, missing-class nodes are completed using statistics from both source and target domains. A global memory bank $\mathcal{M}$ serves as a proxy, and supplementary nodes are generated as $\mathbf{V}_c = \mathcal{M}_c + \mathcal{N}(\mu, \sigma)$, where $\mathcal{N}(\mu, \sigma)$ is a Gaussian with mean $\mu$ from the current-domain memory and standard deviation $\sigma$ from the counterpart domain. The generated nodes $\mathbf{V}_c$ are appended to $\mathbf{S}_\varepsilon$, which is then updated via multi-head self-attention as $\mathbf{S}_\varepsilon \gets \text{Softmax}(\mathbf{S}_\varepsilon W_q (\mathbf{S}_\varepsilon W_k)^\top \sqrt{\frac{1}{d_k}})(\mathbf{S}_\varepsilon W_v) + \mathbf{S}_\varepsilon$ to integrate global dependencies within and across domains. The resulting node set $\mathbf{S}_\varepsilon$, together with edge affinity $\xi = \mathcal{F}_\text{drop} \, \text{Softmax}[\mathbf{S}_\varepsilon W_q (\mathbf{S}_\varepsilon W_k)^\top]$, form the semantic entities and topological structure of the graph, respectively.
\\
\textbf{Graph Matching and Regularization.} This section comprises three loss components. The open-set graph matching loss aligns classes shared across domains, the graph edge affinity loss preserves structural consistency, and the unknown regularization loss encourages separation of unknown classes. From the candidate node set $\mathbf{S}_\varepsilon$, we extract the source and target nodes, $\mathbf{V}_s$ and $\mathbf{V}_t$, respectively. Based on these nodes, we compute a node matching matrix $\mathcal{A} = \mathrm{Sinkhorn}(\mathrm{InstNorm}(\phi(\mathbf{V}_s, \mathbf{V}_t)))$, where $\phi(\cdot, \cdot)$ is a learnable affinity function, $\mathrm{InstNorm}(\cdot)$ normalizes the affinities, and Sinkhorn enforces approximate doubly stochasticity.
Then, we construct the matching labels based on the source domain labels and the target domain pseudo labels. Specifically, we define the open-set matching label while ignoring the unknown class as 
$\mathbf{M} = (\mathbf{H}_s \mathbf{H}_t^\top) \odot ((\mathbf{H}_s \mathbf{e}_{\mathrm{unk}} = 0)(\mathbf{H}_t \mathbf{e}_{\mathrm{unk}} = 0)^\top)$, 
where $\mathbf{H}_s$ and $\mathbf{H}_t$ denote the one-hot matrices of the source and target domain node set, $\mathbf{e}_{\mathrm{unk}}$ is the unit vector corresponding to the unknown class, and $\odot$ represents element-wise multiplication. The graph matching loss for the node set $\mathbf{S}_\varepsilon$ is defined as follows:
\begin{equation}
\small
\begin{aligned}
& \ell_{\mathrm{graph}} = 
\underbrace{
\frac{1}{|\mathbf{S}_\varepsilon|} \sum_{(i,j) \in \mathbf{S}_\varepsilon} (\mathcal{A}_{ij} - \mathbf{M})^2
}_{\text{Graph Matching Loss}}
+
\underbrace{
\frac{1}{|\mathcal{A}|} \big\| \xi_s\mathcal{A}  - \mathcal{A}\xi_t \big\|_1
}_{\text{Graph Edge Loss}} \\
&\quad  
+\underbrace{
\frac{\beta}{|\mathcal{K}_t|\,|\mathcal{U}_t|} \big\| \tilde{\mathbf{V}}_t^{\mathcal{K}_t} (\tilde{\mathbf{V}}_t^{\mathcal{U}_t})^\top \big\|_F^2
+
\frac{\beta}{|\mathcal{K}_s|\,|\mathcal{U}_s|} \big\| \tilde{\mathbf{V}}_s^{\mathcal{K}_s} (\tilde{\mathbf{V}}_s^{\mathcal{U}_s})^\top \big\|_F^2
}_{\text{Unknown-aware Regularization Loss}},
\end{aligned}
\label{graph matching}
\end{equation}
where $\mathcal{K}_{s,t}$ and $\mathcal{U}_{s,t}$ denote the known and unknown node sets, $\xi$ represents the edge affinities, 
$\tilde{\mathbf{V}}$ denotes the unit-normalized node, $\beta$ is the corresponding weight, and 
$\|\cdot\|_1$ and $\|\cdot\|_F$ denote the $\ell_1$ and Frobenius norms, respectively. 
\begin{figure}
    \centering
    \includegraphics[width=1.0\linewidth]{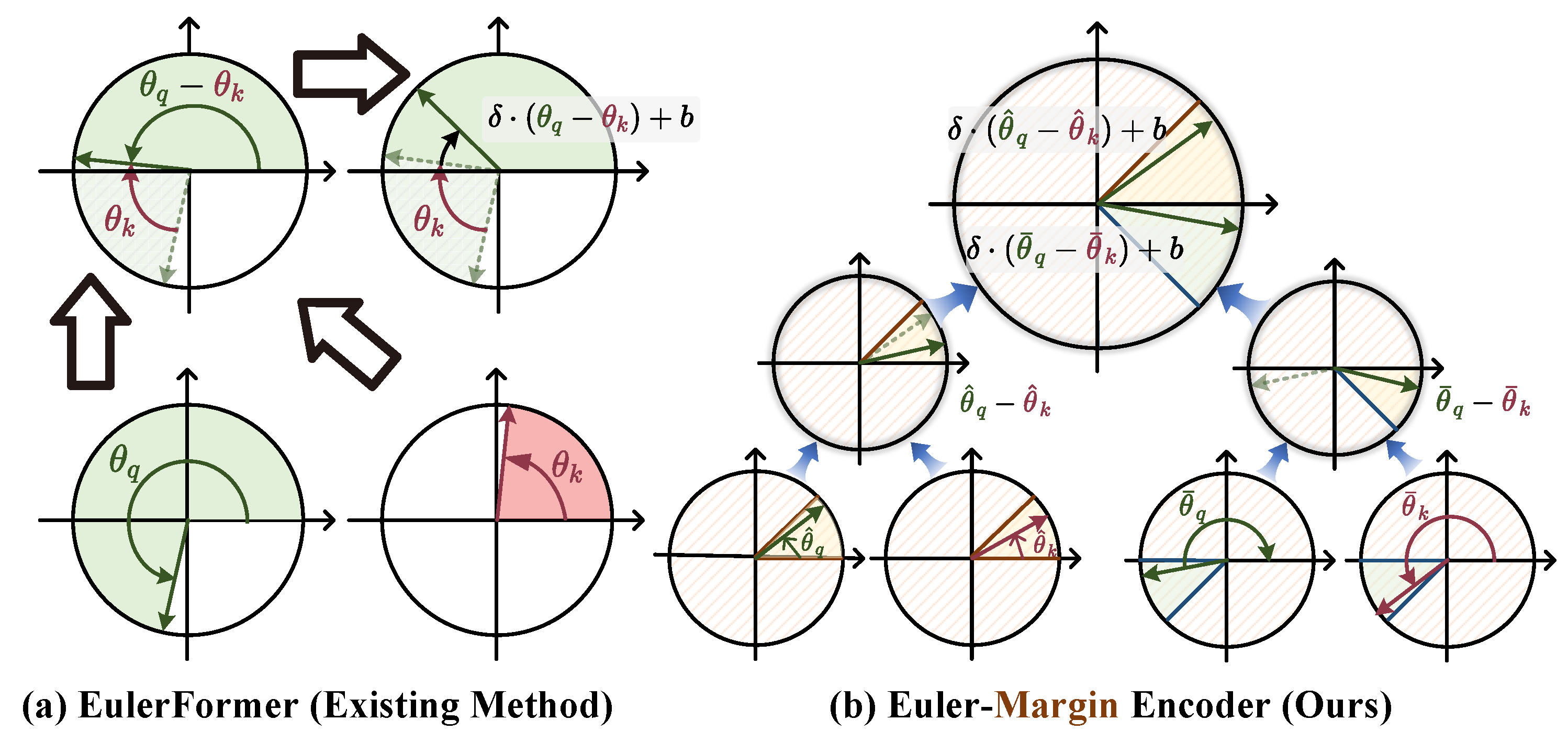}
    \caption{ (a) EulerFormer~\cite{tian2024eulerformer} fails to build a semantic angle constraint, limiting generalization to unseen views. (b) Our method employs Euler-Margin Projection to constrain the angle within the interval, and utilizes Amplitude and Phase Modulation to adjust the class distribution for known and unknown class separation.
}
\label{eulerformer}
\end{figure}
\subsection{Euler-Margin Attention (EMA)}
We propose \textbf{Euler-Margin Attention (EMA)} for semantic extrapolation between base and novel classes under unseen viewpoints. EMA projects channel features into a complex vector space through an angular-margin transformation to alleviate field-of-view distortions and jointly modulates amplitude and phase distributions to enhance class separability.
\\
\textbf{Standard Euler Formula.} We first revisit the Euler formula and define the relevant terms for feature representation following existing works~\cite{su2024roformer, tian2024eulerformer}. Let a feature tensor be $\mathbf{V} \in \mathbb{R}^{B \times N \times d}$, represented in rectangular form as $\mathbf{r} + i \mathbf{s}$, where $\mathbf{r} = \mathbf{V}_{::2}$ denotes the real components corresponding to even channels and $\mathbf{s} = \mathbf{V}_{1::2}$ denotes the imaginary components corresponding to odd channels. The transformation based on Euler’s formula is then applied as follows:
\begin{equation}
\textbf{r} + i\textbf{s}=\Lambda e^{i \theta},\Lambda  = \sqrt{\textbf{r}^2 + \textbf{s}^2}, \quad
\theta = \operatorname{atan2}(\textbf{s},\textbf{r}),
\end{equation}
where $\Lambda$ denotes the magnitude,  $\theta$ is the phase angle, and $\rm atan2(\cdot,\cdot)$ is the arctangent function. As shown in \cref{eulerformer}(a), Existing methods struggle to generalize across views and in open-world scenarios because the angular distribution of the same class shifts across views, and the angles of known and unknown classes overlap. To address these issues, we propose the Euler-Margin Attention (EMA) module, as illustrated in \cref{eulerformer}(b). EMA constrains the angle within a limited range, ensuring intra-class compactness to reduce geometric FoV distortions. Moreover, it introduces learnable parameters to adjust the amplitude (class importance) and phase (class direction) distributions, thereby improving open-set generalization in unseen viewpoints.
\\
\textbf{Euler-Margin Projection.}
Given an input feature tensor $\mathbf{V} \in \mathbb{R}^{B \times N \times d}$, we first apply a channel-wise reordering operator $\pi(\cdot)$, which sorts channels by values in descending order as follows $\pi(\mathbf{V}) = [\mathbf{V}_{1}, \mathbf{V}_{2}, \dots, \mathbf{V}_{d}]$, s.t. $\mathbf{V}_{i+1} < \mathbf{V}_i$. 
The function $\pi(\cdot)$  represents a mapping through the soft permutation matrix, which ensures the backpropagation of the gradient. The reordered channels are then partitioned into real and imaginary components as follows $\operatorname{Re}(\mathbf{V}):= \pi(\mathbf{V})_{::2}$ and $\operatorname{Im}(\mathbf{V}):= \pi(\mathbf{V})_{1::2}$. The Euler-Margin Projection is defined as follows $\mathcal{F}(\mathbf{V}) = \mathcal{F}\left([\pi(\mathbf{V})_{::2}, \pi(\mathbf{V})_{1::2}]\right) = \Lambda \cdot e^{i\theta}$, where $\Lambda$ and $\theta$ denote the amplitude and phase of the complex feature. The channel reorder constrains the angle $\theta$ within the complex vector space, thus mitigating the cross-viewpoint discrepancy. Subsequently, we introduce learnable weights to perform amplitude and phase modulation, thereby adjusting the known class distribution and improving class separation.
\\
\textbf{Amplitude and Phase Modulation.}
We transform the features $\mathbf{V}$ into a complex vector space, resulting in query $\mathbf{V}_q$ and key embeddings $\mathbf{V}_k$. The resulting self-attention dot product is then expressed via the Euler formula as follows:
\begin{equation}
\mathbf{V}_q^\top \mathbf{V}_k = (\Lambda_q \odot \Lambda_k)^\top \mathrm{Re}\!\left[ \exp\big(i(\theta_q - \theta_k)\big) \right],
\end{equation}
where $\Lambda$ and $\theta$ represent the amplitude and phase. Then we introduce learnable scaling and bias factors to modulate both the amplitude and phase components. The resulting modulated attention score, denoted as $\mathcal{E}_{\text{Euler}}$, is   as follows:
\begin{equation} 
\small
\mathcal{E}_{\text{Euler}} = \underbrace{\left(e^{2\delta_1} (\Lambda_q \odot \Lambda_k) \right)^\top}_{\text{Amplitude Modulation}} \mathrm{Re}\left[\exp\left(i \cdot \underbrace{[\delta_2 \cdot (\theta_q - \theta_k) + b]}_{\text{Phase Modulation}}\right)\right],
\label{euler-modualtion}
\end{equation} 
where $\delta_1$ is a learnable exponential weight used to scale the amplitude as $\exp(2\delta_1)$, and $\delta_2$ is a learnable scale for the semantic angle $(\theta_q - \theta_k)$, and $b$ is the bias for the phase. The amplitude characterizes the importance of a feature, while the phase determines its semantic direction. The learnable factors in \cref{euler-modualtion} are optimized for open-set generalization.

\subsection{Model Optimization}
The training objective of our method is defined as follows:
\begin{equation}
\mathcal{L}_{\rm total} = \ell_{\rm seg} + \ell_{\rm mixup} + \gamma \cdot \ell_{\rm graph},
\label{total}
\end{equation}
where $\ell_{\rm seg}$ is the supervised loss on the source domain, $\ell_{\rm mixup}$ is the pseudo-label loss for mixup training between source and target domains, and $\ell_{\rm graph}$ denotes the loss of the proposed GMA module, with $\gamma$ serving as the weight.

\begin{figure}[!t]
    \centering
    \includegraphics[width=0.95\linewidth]{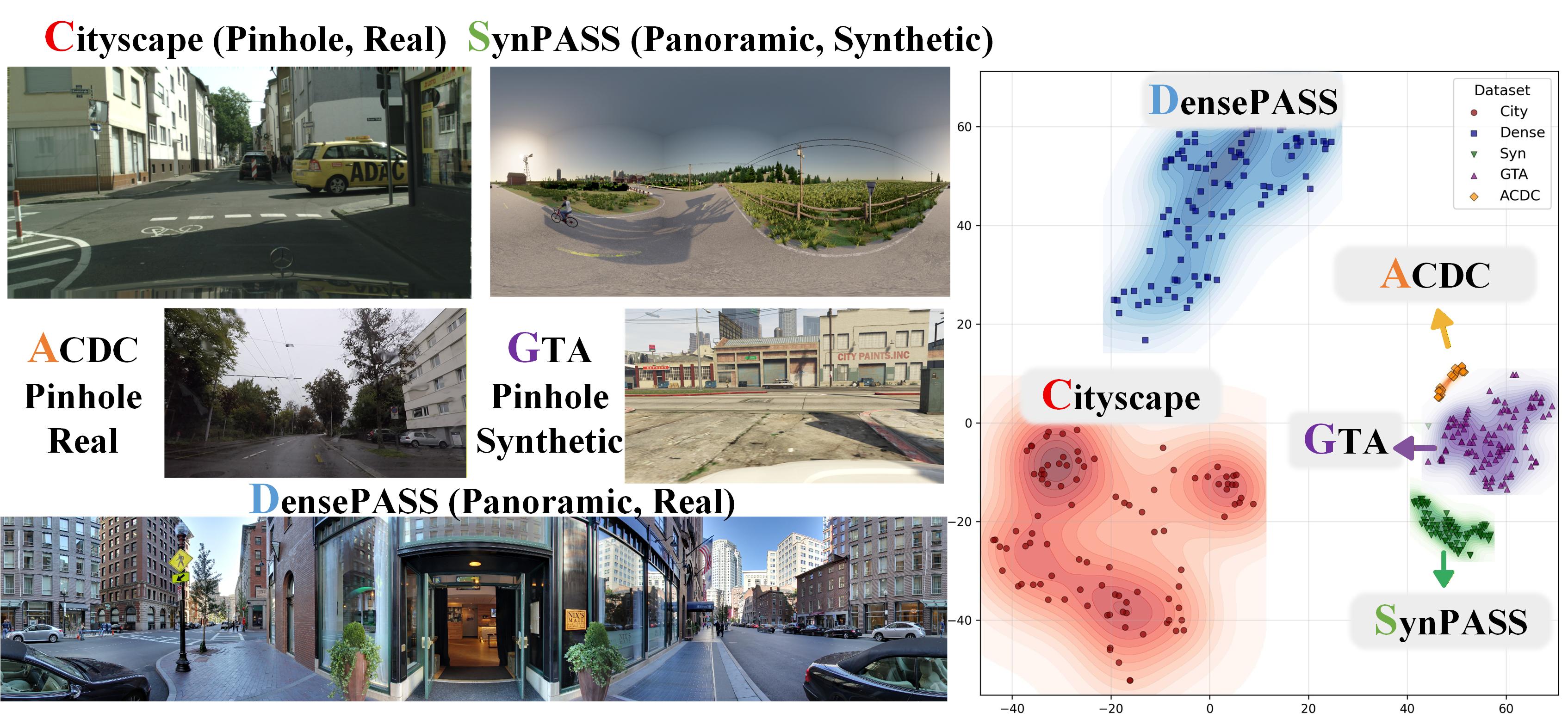}
    \caption{Feature distributions of the car category visualized via t-SNE across multiple datasets, illustrating the effects of  \textit{Camera Types} (Pinhole, Panoramic, Real, and Synthetic) and varying \textit{Weather Conditions} (Fog, Rain, Cloudy, Sunny, Night, and Snow). 
    }
    \label{fig:feature_distributions}
\end{figure}

\definecolor{darkgreen}{RGB}{0,100,0}
\definecolor{darkred}{RGB}{139,0,0}

\begin{table*}[htbp]
\centering
\caption{Comparisons under \textit{\textbf{Camera Shift} \{Pin2Pan, Real2Real\}}  
for Cityscapes $\rightarrow$ DensePASS (\textbf{C2D}) open-set domain adaptation in outdoor scenarios. * indicates an experiment based on DAFormer~\cite{hoyer2022daformer}.} 
\setlength{\tabcolsep}{4pt}
\resizebox{\textwidth}{!}{
\begin{tabular}{l|ccccccccccccc|>{\columncolor{gray!20}}c>{\columncolor{gray!20}}c>{\columncolor{gray!20}}c}
\toprule
Method & Road & S.walk & Build. & Wall & Fence & Tr.light & Veget. & Terrain & Sky & Car & Bus & M.cycle & Bicycle & \textbf{Common} & \textbf{Private} & \textbf{H-Score}  \\
\midrule
OSBP*~\cite{saito2018open} & 39.29 & 24.94 & 84.90 & 34.11 & 40.23 & 27.70 & 78.34 & 16.00 & 26.52 & 53.05 & 21.89 & 51.76 & 34.34 & 41.00 & 7.01 & 11.97 \\
UAN*~\cite{you2019universal}  & 46.78 & 26.59 & 82.61 & 36.57 & 43.02 & 18.97 & 79.09 & 21.09 & 43.51 & 65.06 & 13.30 & 48.97 & 20.31 & 41.99 & 0.93 & 1.81 \\
UniOT*\cite{jang2022unknown} & 63.59 & 31.37 & 86.79 & 42.35 & 38.04 & 18.90 & 79.82 & 37.86 & 81.45 & 68.89 & 37.91 & 59.10 & 31.67 & 52.13 & 0.00 &  0.00 \\
\midrule
DAFormer~\cite{hoyer2022daformer} & 65.37 & 34.95 & 85.75 & 35.89 & 37.61 & 30.46 & 79.16 & 36.36 & 84.46 & 69.77 & 31.61 & 62.26 & 37.07 & 53.13 & 0.00 & 0.00 \\
MIC~\cite{hoyer2023mic}  & 51.04 & 15.66 & 84.28 & 36.06 & 35.34 & 30.55 & 76.44 & 19.56 & 73.52 & 72.81 & 28.31 & 55.48 & 29.70 & 46.83 & 0.00 & 0.00 \\
HRDA~\cite{hoyer2022hrda}   & 68.84 & 34.89 & 85.71 & 38.39 & 39.64 & 28.12 & 79.58 & 38.74 & 90.78 & 
68.23 & 29.37 & 59.62 & 32.53 & 53.42 & 0.00 & 0.00    \\
BUS (SAM)~\cite{choe2024open} & 60.63 & 28.46 & 82.04 & 29.68 & 32.47 & 26.11 & 77.73 & 24.96 & 94.46 & 68.02 & 39.87 & 43.37 & 35.38 & 49.47 & 3.10 & 5.84 \\

\rowcolor{blue!15} %
\midrule
Ours  (SAM) & 74.86 & 34.20 & 85.90 & 40.40 & 43.52 & 31.04 & 79.87 & 41.27 & 94.84 & 78.85 & 32.83 & 63.92 & 37.06 & 56.81 & 18.86 & 28.32   \\
\bottomrule
\end{tabular}
}
\label{c2d}
\end{table*}

\begin{table*}[htbp]
\centering
\caption{Comparisons under  \textit{\textbf{Camera Shift} \{Syn2Real, Pan2Pan\}} 
for SynPASS $\rightarrow$ DensePASS  (\textbf{S2D})  open-set domain adaptation in outdoor scenarios. * indicates an experiment based on DAFormer~\cite{hoyer2022daformer}.}
\setlength{\tabcolsep}{4pt}
\resizebox{\textwidth}{!}{
\begin{tabular}{l|ccccccccccccc|>{\columncolor{gray!20}}c>{\columncolor{gray!20}}c>{\columncolor{gray!20}}c}
\toprule
Method & Road & S.walk & Build. & Wall & Fence & Pole & Tr.light & Tr.sign & Veget. & Terrain & Sky & Person & Car & \textbf{Common} & \textbf{Private} & \textbf{H-Score}   \\
\midrule
OSBP*~\cite{saito2018open} & 34.68 & 0.20 & 23.01 & 3.17 & 2.37 & 0.74 & 0.00 & 0.27 & 25.58 & 4.69 & 56.97 & 0.17 & 3.91 & 11.98 & 5.77 & 7.79 \\
UAN*~\cite{you2019universal} & 51.01 & 2.22 & 33.48 & 1.52 & 7.33 & 7.73 & 0.00 & 0.59 & 29.75 & 12.07 & 58.77 & 10.0 & 6.72 & 17.01 & 0.83 & 1.59 \\
UniOT*\cite{jang2022unknown} & 47.14 & 2.73 & 32.15 & 1.64 & 5.12 & 7.93 & 0.00 & 0.20 & 27.71 & 1.01 & 50.80 & 21.97 & 2.24 & 15.44 & 0.00 & 0.00 \\
\midrule
DAFormer~\cite{hoyer2022daformer} & 52.20 & 4.44 & 34.42 & 3.24 & 10.16 & 5.80 & 0.00 & 0.12 & 56.07 & 6.33 & 63.67 & 37.26 & 2.31 & 21.23 & 0.00 & 0.00 \\ 
MIC~\cite{hoyer2023mic}   & 48.23 & 3.6 & 46.16 & 0.04 & 10.52 & 4.77 & 0.00 & 0.20 & 25.68 & 3.52 & 65.18 & 3.27 & 2.83 & 16.46 & 0.00 & 0.00 \\
HRDA~\cite{hoyer2022hrda}    &  53.18 & 7.29 & 36.77 & 3.04 & 9.35 & 5.09 & 0.00 & 0.09 & 46.23 & 
5.41 & 61.59 & 17.58 & 3.09 & 19.13 & 0.00 & 0.00    \\
BUS (SAM)~\cite{choe2024open} & 42.27 & 4.27 & 12.06 & 0.05 & 8.40 & 10.07 & 0.00 & 0.00 & 0.79 & 1.68 & 80.69 & 0.58 & 0.22 & 12.39 & 1.73 & 3.04 \\
\midrule
\rowcolor{blue!15} %
Ours  (SAM) & 70.19 & 35.28 & 58.34 & 8.02 & 0.68 & 20.7 & 0.00 & 0.57 & 75.39 & 30.67 & 92.27 & 45.86 & 17.95 & 35.07 & 7.48 & 12.33 \\
\bottomrule
\end{tabular}
}
\label{s2d}
\end{table*}

\begin{table*}[htbp]
\centering
\caption{Comparisons under different  \textit{\textbf{Camera Shift} (Pin2Pan,Syn2Syn }) 
and \textit{\textbf{Weather Shift} \{\textcolor{darkgreen}{Sunny}\}$\rightarrow$\textcolor{darkgreen}{\{Sunny}, \textcolor{darkred}{Cloudy, Fog, Rain, Night}\}} for \textit{GTA $\rightarrow$ SynPASS}  (\textbf{G2S}) open-set domain adaptation in outdoor scenarios. * indicates an experiment based on DAFormer~\cite{hoyer2022daformer}.}
\setlength{\tabcolsep}{4pt}
\resizebox{\textwidth}{!}{
\begin{tabular}{l|ccccccccccccc|>{\columncolor{gray!20}}c>{\columncolor{gray!20}}c>{\columncolor{gray!20}}c}
\toprule
Method & Road & S.walk & Build. & Wall & Fence & Pole & Tr.light & Tr.sign & Veget. & Terrain & Sky & Person & Car & \textbf{Common} & \textbf{Private} & \textbf{H-Score}   \\
\midrule
OSBP*~\cite{saito2018open} & 75.30 & 35.06 & 40.29 & 0.00 & 11.27 & 15.14 & 8.60 & 21.40 & 43.69 & 47.07 & 57.05 & 32.83 & 43.37 & 33.16 & 0.00 & 0.00 \\
UAN*~\cite{you2019universal} & 67.23 & 26.45 & 31.54 & 0.06 & 6.25 & 19.58 & 20.71 & 18.37 & 44.69 & 39.72 & 27.94 & 22.41 & 25.22 & 26.94 & 0.73 & 1.42 \\
UniOT*\cite{jang2022unknown} & 88.10 & 57.96 & 56.18 & 0.07 & 14.98 & 19.95 & 18.89 & 14.41 & 42.46 & 47.94 & 87.44 & 16.95 & 61.36 & 40.52 & 0.00 & 0.00 \\
\midrule
DAFormer~\cite{hoyer2022daformer} & 95.69 & 51.33 & 58.01 & 0.02 & 9.67 & 22.30 & 8.77 & 14.88 & 53.18 & 43.83 & 88.08 & 32.80 & 64.98 & 41.81 & 0.00 & 0.00  \\  
MIC~\cite{hoyer2023mic}  & 97.38 & 59.72 & 53.17 & 0.52 & 11.56 & 20.57 & 24.20 & 19.46 & 56.61 & 44.62 & 84.15 & 28.69 & 64.90 & 43.51 & 0.00 & 0.00  \\
HRDA~\cite{hoyer2022hrda}    & 91.13 & 54.68 & 49.92 & 0.24 & 10.33 & 20.03 & 20.70 & 13.25 & 41.23 & 37.74 & 80.60 & 29.12 & 67.33 & 39.72 & 0.00 & 0.00 \\
BUS (SAM)~\cite{choe2024open} & 95.72 & 0.30 & 74.25 & 0.98 & 0.14 & 20.37 & 21.78 & 9.65 & 54.91 & 44.92 & 96.27 & 28.08 & 72.51 & 39.99 & 7.97 & 13.29 \\
\midrule
\rowcolor{blue!15} %
Ours  (SAM) & 94.50 & 50.42 & 73.20 & 0.98 & 20.97 & 15.86 & 12.07 & 17.11 & 54.33 & 46.84 & 96.14 & 27.75 & 74.32 & 44.96 & 10.20 & 16.63 \\
\bottomrule
\end{tabular}
}
\label{g2s}
\end{table*}

\begin{table*}[htbp]
\centering
\caption{Comparisons under  \textit{\textbf{Camera Shift} (Syn2Real, Pan2Pin})  
and \textit{\textbf{Weather Shift} \{\textcolor{darkgreen}{Fog, Rain, Night, Sunny}, \textcolor{darkred}{Cloudy}\}$\rightarrow$ \{\textcolor{darkgreen}{Fog, Rain, Night, Sunny}, \textcolor{darkred}{Snow}}\}  
for SynPASS$\rightarrow$ACDC  (\textbf{S2A}) open-set domain adaptation in outdoor scenarios. * indicates DAFormer-based results~\cite{hoyer2022daformer}.}
\label{tab:pin2pan_selected}
\setlength{\tabcolsep}{4pt}
\resizebox{\textwidth}{!}{
\begin{tabular}{l|ccccccccccccc|>{\columncolor{gray!20}}c>{\columncolor{gray!20}}c>{\columncolor{gray!20}}c}
\toprule
Method & Road & S.walk & Build. & Wall & Fence & Pole & Tr.light & Tr.sign & Veget. & Terrain & Sky & Person & Car & \textbf{Common} & \textbf{Private} & \textbf{H-Score}   \\
\midrule
OSBP*~\cite{saito2018open}       & 52.79 & 0.89 & 6.72 & 0.80 & 0.41 & 0.04 & 0.00 & 0.00 & 32.79 & 0.00 & 73.7 & 0.21 & 0.14 & 12.96 & 5.02 & 7.24 \\
UAN*~\cite{you2019universal} & 67.50 & 7.27 & 26.60 & 0.97 & 1.26 & 14.20 & 0.00 & 0.00 & 50.96 & 13.10 & 67.25 & 0.23 & 0.34 & 19.21 & 0.13 & 0.26 \\
UniOT*\cite{jang2022unknown} & 57.98 & 2.00 & 20.36 & 0.03 & 2.09 & 4.51 & 0.00 & 0.00 & 59.15 & 0.51 & 87.66 & 8.53 & 4.52 & 19.03 & 0.00 & 0.00 \\
\midrule
MIC~\cite{hoyer2023mic}    & 62.32 & 10.60 & 60.77 & 0.00 & 1.17 & 3.18 & 0.00 & 0.00 & 60.02 & 4.71 & 80.46 & 26.67 & 11.82 & 24.75 & 0.00 & 0.00 \\
DAFormer~\cite{hoyer2022daformer} & 66.49 & 6.90 & 30.34 & 0.07 & 3.12 & 1.96 & 0.00 & 0.00 & 75.47 & 19.46 & 93.97 & 33.34 & 0.01 & 25.47 & 0.00 & 0.00 \\ 
HRDA~\cite{hoyer2022hrda}   & 66.40 & 8.37 & 31.28 & 2.05 & 1.66 & 5.77 & 0.00 & 0.00 & 57.23 & 19.00 & 75.79 & 37.27 & 14.81 & 24.59 & 0.00 & 0.00 \\
BUS  (SAM)~\cite{choe2024open} & 71.86 & 10.95 & 16.77 & 0.00 & 0.03 & 0.03 & 0.00 & 0.00 & 17.25 & 9.54 & 82.61 & 0.00 & 0.00 & 16.08 & 3.70 & 6.01 \\
\midrule
\rowcolor{blue!15} %
Ours  (SAM) & 52.53 & 9.24  & 54.58  & 10.4  & 0.28  & 24.39 & 0.00     & 0.00     & 52.65  & 18.95  & 83.16 & 32.05  & 54.04  & 30.17   & 9.18  & 14.08 \\
\bottomrule
\end{tabular}
}
\label{s2a}
\end{table*}

\section{Experiments}
\label{sec:experiment}
\subsection{Benchmark Setup}

\noindent\textbf{Datasets.} 
We conduct comprehensive experiments on multiple datasets to evaluate the proposed method under both geometric and semantic domain shifts. 
\textbf{DensePASS}~\cite{ma2021densepass} is a real-world panoramic dataset covering diverse city scenes collected with panoramic sensors. It provides $2{,}000$ unlabeled images for transfer optimization and $100$ labeled images for evaluation. 
\textbf{SynPASS}~\cite{zhang2024behind} is a synthetic panoramic dataset rendered under various weather conditions, including \textit{Cloudy}, \textit{Foggy}, \textit{Rainy}, \textit{Sunny}, and \textit{Night}, containing $9{,}080$ panoramic images in total. 
For pinhole-domain studies, we adopt three widely used benchmarks. 
\textbf{Cityscapes}~\cite{cordts2016cityscapes} is a real-world dataset with $2{,}975$ training and $500$ validation images annotated with $19$ semantic categories. 
\textbf{GTA5}~\cite{richter2016playing} is a large-scale synthetic dataset consisting of $24{,}966$ images generated from the GTA-V game engine, providing pixel-level labels consistent with Cityscapes. 
\textbf{ACDC}~\cite{sakaridis2021acdc} is a real-world data set that emphasizes adverse weather (\textit{Foggy}, \textit{Night}, \textit{Rainy}, and \textit{Snowy}), including $1{,}600$ training and $406$ validation images. 

\noindent\textbf{Benchmark Settings.}
Our domain adaptation evaluation encompasses two complementary scenarios. Camera Shift addresses transfer between pinhole and panoramic images ($\text{Pin}\leftrightarrow\text{Pan}$) and between synthetic and real domains ($\text{Syn}\rightarrow\text{Real}$). Weather Shift investigates adaptation across six diverse conditions: $\text{Fog}$, $\text{Rain}$, $\text{Cloudy}$, $\text{Sunny}$, $\text{Night}$, and $\text{Snow}$. Furthermore, we assess each setting in an open-set configuration where the common class ratio controls the semantic overlap (\textit{e.g.}, $\text{C2D}:68.4\%$, $\text{S2D}:48.1\%$, $\text{G2S}:48.1\%$, $\text{S2A}:48.1\%$). As shown in \cref{fig:feature_distributions}, we analyze the domain gap among the datasets. Real datasets like Cityscapes~\cite{cordts2016cityscapes} and DensePASS~\cite{ma2021densepass} exhibit higher variance than synthetic datasets (GTA~\cite{richter2016playing}, SynPASS~\cite{zhang2024behind}), while the adverse weather dataset ACDC~\cite{sakaridis2021acdc} has weaker class representation compared to the real datasets. The supplementary material provides complete details of the benchmark protocols, dataset splits, and category settings.
\\
\noindent\textbf{Evaluation Metrics.}
We adopt the mean Intersection-over-Union (mIoU) to evaluate segmentation quality for both base and novel classes for {Common} and {Private}. 
In addition, we adopt the {H-Score} between the mIoU of the base classes and the IoU of novel classes, following the common open-set evaluation practice.  This metric jointly reflects the performance on both shared and novel semantic spaces.
\\
\noindent\textbf{Implementation Details.}
Our implementation follows the standard configuration of DAFormer~\cite{hoyer2022daformer}. 
We employ MiT-B5~\cite{xie2021segformer} as the backbone, initialized with ImageNet-1K pretrained weights. 
All models are trained for $40k$ iterations using $512{\times}512$ random crops on both the source (pinhole) and target (panoramic) domains. 
During testing, panoramic images are evaluated at their original resolution. 
To improve pseudo-label quality in the target domain, we utilize MobileSAM~\cite{zhang2023faster} for mask refinement, following the same refinement strategy as in the prior Open-Set UDA method~\cite{choe2024open}.

\subsection{Performance Analysis across Benchmarks}

\textbf{C2D.} As shown in \cref{c2d}, we report the performance results for the Cityscapes$\rightarrow$DensePASS under the \textit{Camera Shift} setting. The results indicate that our proposed method achieves higher mIoU on both common and private categories, suggesting that the proposed method can distinguish between known and unknown categories. Additionally, the observed improvements in the H-score imply an enhanced capability to identify private categories. These results also suggest that, when transferring from pinhole to panoramic images, the GMA module better captures inter-category relationships while remaining robust to geometric distortions.
\\
\textbf{S2D.}
As shown in \cref{s2d}, the open-set performance on SynPASS$\rightarrow$DensePASS under \textit{Camera Shift} reflects the challenging nature of synthetic-to-real transfer. Our method consistently outperforms existing approaches on important object categories such as Person and Car, clearly demonstrating that variations and distortions inherent in different panoramic datasets can significantly degrade the quality of object class representations. The proposed GMA and EMA modules are specifically designed to address these distortions and are shown to mitigate their negative impact. Compared to the existing SAM-based method (BUS), our proposed method demonstrates significant advantages in both mIoU results of both common and private categories.
\\
\textbf{G2S.} As shown in \cref{g2s}, the open-set adaptation results for GTA$\rightarrow$SynPASS  under \textit{Camera Shift} and \textit{Weather Shift} highlight the difficulty of transferring from synthetic pinhole to panoramic real-world domains. Our method outperforms the second-best approach by 3.15\% in mIoU, demonstrating that the proposed EMA and GMA modules mitigate image distortions within the same simulation scenario and adverse weather conditions.  Our method performs comparably to existing open-set methods for the Wall class.
\\
\textbf{S2A.} As shown in \cref{s2a}, the open-set adaptation results for SynPASS$\rightarrow$ACDC under the combined \textit{Camera Shift} and \textit{Weather Shift} settings demonstrate model performance in diverse conditions, encompassing both adverse and clear scenes. Despite variations in weather and viewing angles, our method consistently and significantly outperforms existing approaches, particularly in recognizing unknown categories. Specifically, it achieves a 4.16\% mIoU improvement on private and a 6.84\% H-Score increase over the OSBP method, indicating that the proposed module generalizes robustly under diverse weather conditions and FoV distortions. Moreover, in the Pole category, our method delivers markedly superior results compared to existing open-set and closed-set methods overall and consistently.
\subsection{Ablation Study}
\textbf{Module Ablation.}
As shown in \cref{Ablation}, we conduct an ablation study to evaluate the contributions of the GMA and EMA modules in our framework.  Exp.\ding{172}, the baseline without either module, achieves the lowest performance, indicating the necessity of both components. Introducing GMA alone (Exp.\ding{173}) improves both common and private feature representations, demonstrating its effectiveness in aligning shared semantics under FoV shifts. Similarly, incorporating EMA alone (Exp.\ding{174}) enhances viewpoint-invariant representation, boosting overall segmentation performance.
\\
\textbf{Ablation analysis of GMA.} As shown in \cref{gma}, we ablate the GMA loss components of unknown-class learning, graph matching, and affinity, defined in \cref{graph matching}. Removing the graph matching term notably degrades performance, confirming its importance, while the unknown-class component further improves the private class mIoU and H-Score.
\\
\textbf{Sensitivity of  GMA.} As shown in \cref{Sensitivity}, we analyze the sensitivity of the loss weight $\gamma$. The results indicate that $\gamma = 1.0$ favors the private class mIoU and improves the H-score but negatively impacts the performance on common classes mIoU, whereas $\gamma = 0.05$ has the opposite effect. To achieve a balanced trade-off, we set $\gamma = 0.1$ in \cref{total}.
\\
\textbf{Effectiveness of EMA.}
As shown in \cref{ema_eff}, we evaluate the effectiveness of the EMA module against the attention in panoramic segmentation~\cite{zhang2024behind} and Eulerformer~\cite{tian2024eulerformer}. Our method consistently outperforms these baselines in both common and private class mIoU, with a particularly large improvement in H-Score, demonstrating its superior ability in novel class discovery and closed-set class classification.
\begin{figure}[!t]
    \centering
    \includegraphics[width=1.0\linewidth]{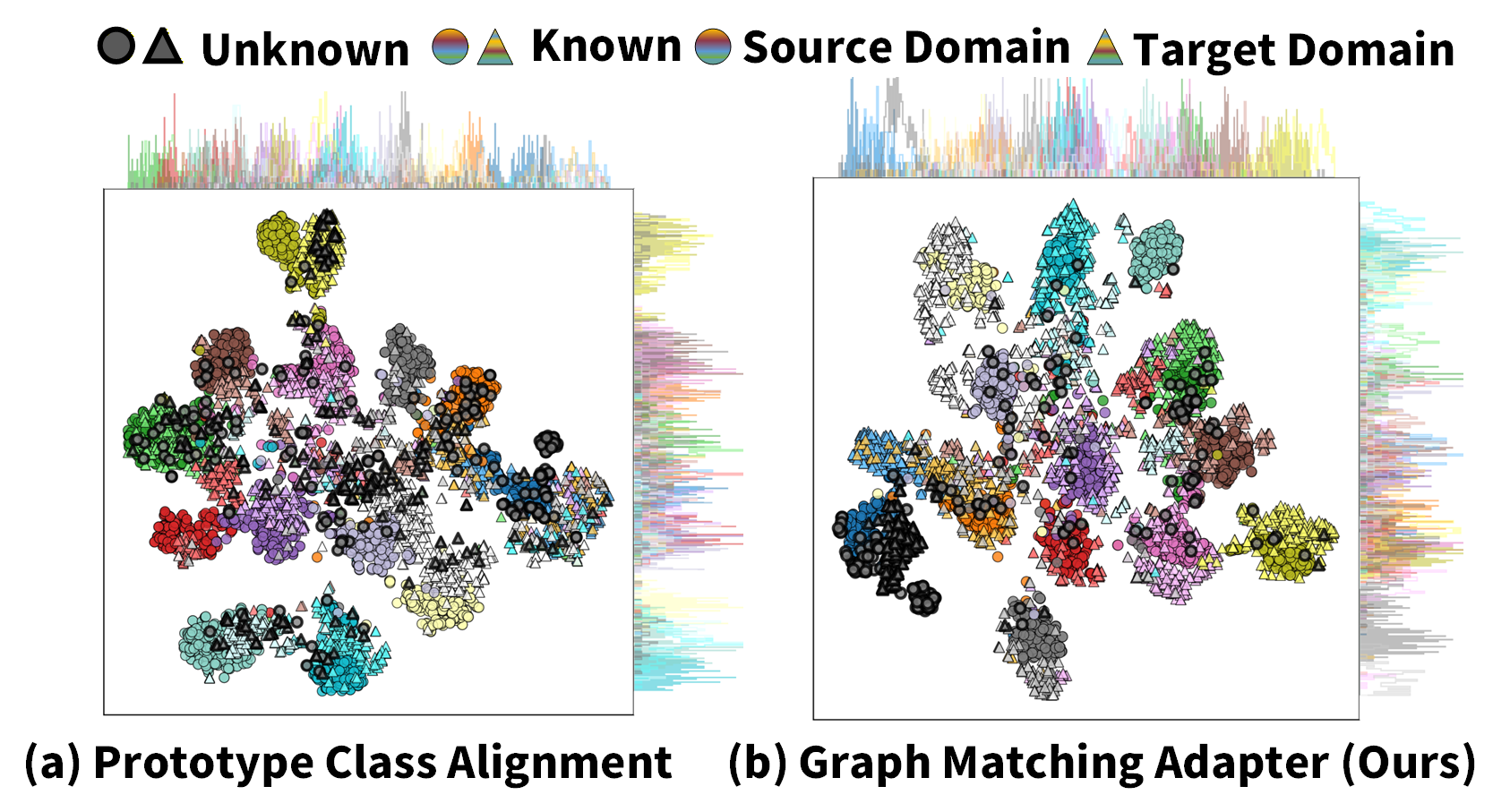}
\caption{\textbf{T-SNE visualization} of source and target domains in \textbf{EDA-PS}eg. (a) The prototype exhibits a relatively mixed unknown class. (b) Ours is the relatively separable unknown class.}
    \label{fig:prototype}
\end{figure}
\\
\textbf{Domain Alignment Analysis.} As illustrated in \cref{fig:prototype}, we visualize the prototypes and GMA features using T-SNE~\cite{maaten2008visualizing}. For each class, features are randomly sampled, where circles represent the source domain and triangles denote the target domain. The accompanying histograms clearly show the sample frequency across bin intervals, with distinct colors corresponding to different categories. Compared with prototype class alignment, GMA consistently exhibits a more distinct separation of unknown classes.

\begin{table}[t]
\centering
\caption{\textbf{Ablation study} of the components in our framework.}
\resizebox{\linewidth}{!}{
\begin{tabular}{c|c|c|c|c|c}
\hline
\multicolumn{6}{c}{Cityscapes $\rightarrow$ DensePASS (\textbf{C2D})} \\ %
\hline
\rowcolor{gray!10}
\textbf{Exp.} & \textbf{GMA} & \textbf{EMA} & \textbf{Common} & \textbf{Private} & \textbf{H-Score} \\
\hline
\ding{172} & - & - & 52.56 & 8.57 & 14.74 \\
\ding{173} & \cmark &  & 55.15  & 14.67 & 23.18   \\
\ding{174} &  & \cmark  &56.12   & 13.00 & 21.11  \\
\hline
\rowcolor{blue!15}
\ding{175} & \cmark & \cmark &  56.81 & 18.86 & 28.32  \\
\hline
\end{tabular}
}
\label{Ablation}
\end{table}

\begin{table}[t]
\centering
\caption{\textbf{Ablation analysis of the GMA module.}}
\resizebox{\linewidth}{!}{
\begin{tabular}{c|l|c|c|c} %
\hline
\multicolumn{5}{c}{Cityscapes $\rightarrow$ DensePASS (\textbf{C2D})} \\ %
\hline
\rowcolor{gray!10}
\textbf{Exp.} & \textbf{Method} & \textbf{Common} & \textbf{Private} & \textbf{H-Score} \\
\hline
\ding{172}  & GMA w/o Unknown & 54.72 & 7.78 & 13.62 \\
\ding{173}  & GMA w/o Matching & 52.75 & 4.76 & 8.73 \\
\ding{174}  & GMA w/o Affinity & 54.06 & 11.59 & 19.09 \\
\hline
\rowcolor{blue!15}
\ding{175}  & GMA (Full) & 55.15 & 14.67 & 23.18 \\
\hline
\end{tabular}
}
\label{gma}
\end{table}

\begin{table}[t]
\centering
\caption{\textbf{Sensitivity of the GMA module loss to the  factor $\gamma$.}}
\resizebox{\linewidth}{!}{
\begin{tabular}{c|c|c|c|c} 
\hline
\multicolumn{5}{c}{Cityscapes $\rightarrow$ DensePASS (\textbf{C2D})} \\
\hline
\rowcolor{gray!10}
\textbf{Exp.} & \centering \textbf{Weight $\gamma$} & \textbf{Common} & \textbf{Private} & \textbf{H-Score} \\
\hline
\ding{172} & -    & 52.56 & 8.57 & 14.74 \\
\ding{173} & 1.00   & 47.16  & 15.45 &  23.28 \\
\ding{174} & 0.10   & 55.15  & 14.67 & 23.18  \\
\ding{175} & 0.05  & 54.57 & 7.90  & 13.80  \\
\ding{176} & 0.01  & 51.72  & 4.59 & 8.43 \\
\hline
\end{tabular}
}
\label{Sensitivity}
\end{table}

\begin{table}[t]
\centering
\caption{\textbf{Effectiveness analysis of the EMA module, added at the same position as in the baseline with the same layers.}}
\resizebox{\linewidth}{!}{
\begin{tabular}{c|l|c|c|c} %
\hline
\multicolumn{5}{c}{Cityscapes $\rightarrow$ DensePASS (\textbf{C2D})} \\ %
\hline
\rowcolor{gray!10}
\textbf{Exp.} & \textbf{Method} & \textbf{Common} & \textbf{Private} & \textbf{H-Score} \\
\hline
\ding{172}  & Baseline & 52.56 & 8.57 & 14.74 \\
\ding{173}  & Self-Attention~\cite{vaswani2017attention} & 55.45 & 10.95 & 18.28 \\
\ding{174}  & Eulerformer~\cite{tian2024eulerformer} & 55.09 & 7.20 & 12.74 \\
\ding{175}  & Deformable MLP~\cite{zhang2024behind} & 55.89 & 7.68 & 13.51   \\
\hline
\rowcolor{blue!15}
\ding{176}  & Euler-Margin (Ours) & 56.12 & 13.0 & 21.11 \\
\hline
\end{tabular}}
\label{ema_eff}
\end{table}

\section{Conclusion}
In this work, we investigate a novel and challenging task setting and propose the \textbf{E}xtrapolative \textbf{D}omain \textbf{A}daptive \textbf{P}anoramic \textbf{S}egmentation (\textbf{E}\textbf{D}\textbf{A}-\textbf{P}\textbf{S}eg). \textbf{E}\textbf{D}\textbf{A}-\textbf{P}\textbf{S}eg aims to enable local-crop transfer learning from limited pinhole viewpoints to full $360^{\circ}$ panoramic scenes while simultaneously discovering novel classes. To address the challenges of unseen-viewpoint open-set class learning and field-of-view distribution discrepancies, we introduce two key components, Euler-Margin Attention (EMA) and the Graph Matching Adapter (GMA). Extensive experiments on the $\textit{Camera Shift}$ and $\textit{Weather Shift}$ benchmarks demonstrate that our approach achieves state-of-the-art performance across diverse resolutions and viewing geometries.
\\
\noindent\textbf{Limitations.}  Random cropping introduces sampling sensitivity, occasionally leading to slight training instability. Graph matching increases both model parameters and computational cost, while the Euler-Margin Attention mechanism adds further architectural complexity. Future work will explore strategies to improve efficiency and stability.
\clearpage
\section*{Acknowledgment}

This work was supported in part by the National Natural Science Foundation of China (Grant No. 62473139), in part by the Hunan Provincial Research and Development Project (Grant No. 2025QK3019), in part by the State Key Laboratory of Autonomous Intelligent Unmanned Systems (the opening project number ZZKF2025-2-10), and in part by the Deutsche Forschungsgemeinschaft (DFG, German Research Foundation) - SFB 1574 - 471687386. This research was partially funded by the Ministry of Education and Science of Bulgaria (support for INSAIT, part of the Bulgarian National Roadmap for Research Infrastructure).
{
    \small
     \bibliographystyle{ieeenat_fullname}
     \bibliography{main}
}

\clearpage

\newcounter{supsec}
\newcommand{\SupSection}[1]{%
    \stepcounter{supsec}%
    \section*{Sec. \Alph{supsec}: #1}%
    \renewcommand{\thesection}{\Alph{supsec}}%
    \renewcommand{\thesubsection}{\Alph{supsec}.\arabic{subsection}}%
    \setcounter{subsection}{0}%
}

\begin{figure*}[!t]
    \centering
    \includegraphics[width=1.0\linewidth]{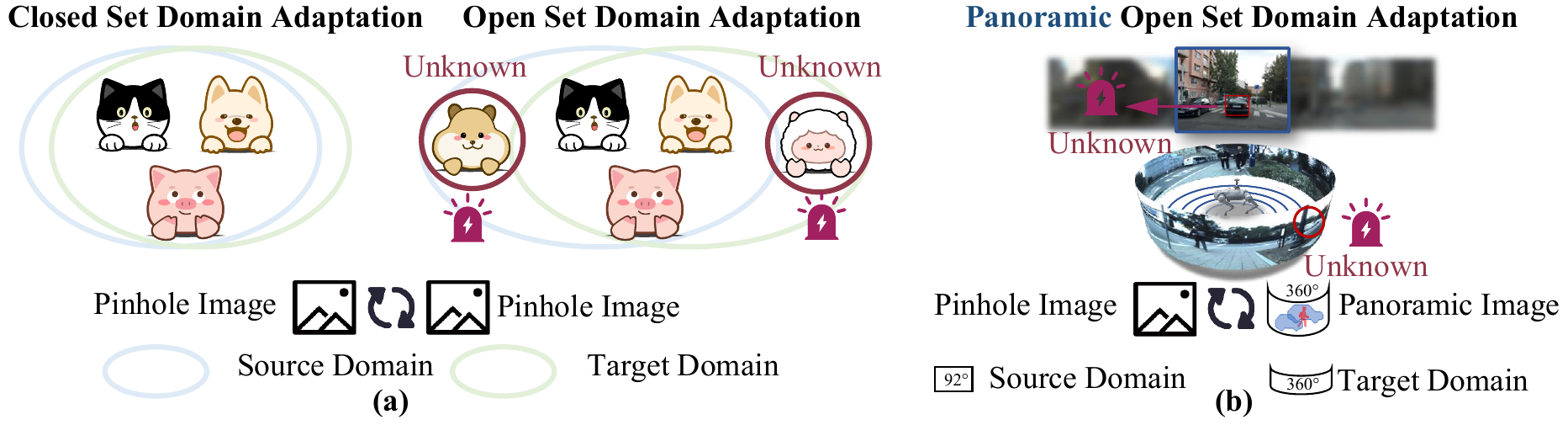}
    \caption{Conceptual comparison of the domain adaptation setting.}
    \label{fig:closed_set}
\end{figure*}

In this supplementary material, we provide three sections to complement the main manuscript. 
\textbf{Section \color{red}{A}} offers a detailed description of the proposed task setting and compares it with related settings. 
\textbf{Section \color{red}{B}} presents additional experiments to demonstrate the effectiveness of the proposed module and includes an analysis of its sensitivity coefficients.
\textbf{Section \color{red}{C}} discusses the limitations of our method and provides an outlook on the societal implications.

\section*{Sec. \textcolor{red}{A}: Clarification and Discussion}

\begin{itemize}
\item {Task Clarification}
\item {Benchmark Setup}
\item {Implementation Details}
\end{itemize}

\section*{Sec. \textcolor{red}{B}: Quantitative Comparison}

\begin{itemize}
    \item {Further Analysis}
    \item {Sensitivity Analysis}
    \item {Model Efficiency}
    \item {Visualization Analysis}
\end{itemize}

\section*{Sec. \textcolor{red}{C}: Limitations and Outlook}

\begin{itemize}
    \item Societal Implications
    \item Future Research Directions
    \item Limitations and Potential Solutions
\end{itemize}

\SupSection{Clarification and Discussion}
\subsection{Task Clarification}
\textbf{Clarification of the setting.} As illustrated in Fig.~\ref{fig:closed_set}, the figure depicts the conceptual differences among three domain adaptation settings: Closed-Set Domain Adaptation, Open-Set Domain Adaptation, and Panoramic Open-Set Domain Adaptation.
(a) \textbf{Closed-Set Domain Adaptation}: The source domain and the target domain share an identical set of classes (for example, cats and dogs). The objective is to reduce the domain discrepancy under the assumption that all labels are fully shared.
(b) \textbf{Open-Set Domain Adaptation}: The target domain contains additional unknown categories that do not appear in the source domain. The goal is to properly align the shared classes while identifying and filtering out target samples belonging to unknown categories.
(c) \textbf{\color{purple}Panoramic Open-Set Domain Adaptation (PODA)}: This setting extends the open-set scenario to a cross-modal context, where the source domain provides pinhole images and the target domain provides panoramic images. The target domain has known and unknown classes, causing challenges due to camera and semantic differences.
\\
\textbf{Clarification of the domain shift.}  From Table~\ref{comparing setting}, it is clear that conventional Unsupervised Domain Adaptation (UDA) methods are unable to handle open-set categories or adapt to diverse weather conditions. While Panoramic UDA extends standard UDA by explicitly modeling variations in the field of view (FoV), it still cannot handle unknown categories or environmental shifts. Open-Set Domain Adaptation focuses exclusively on unseen classes, and Open Compound Domain Adaptation aims to handle domain shifts caused by varying weather conditions. However, both neglect the FoV discrepancies inherent to panoramic imagery. In contrast, the proposed \textbf{\color{purple} PODA}  framework simultaneously addresses open-set recognition, weather-related domain shifts, and FoV variations. By jointly modeling these three critical factors, {\color{purple} \textbf{PODA}} achieves superior generalization and robustness, enabling more reliable cross-domain perception in complex real-world driving scenarios.
\subsection{Technical Clarifications}
\subsection{Benchmark Setup}
\textbf{Dataset and class configuration.}
 To evaluate open-set cross-domain semantic segmentation in a controlled yet diverse manner, we construct a benchmark spanning real-to-real, synthetic-to-real, and synthetic-to-synthetic transfers. For source and target domains, we conduct the \textbf{domain-shared (base)} and \textbf{domain-specific (private)} classes, where the latter appear only in one domain and thus represent unforeseen categories encountered during deployment. Weather conditions follow the same convention, capturing whether environmental factors are shared or exclusive to a single domain. The benchmark includes four representative transfers as CityScapes$\rightarrow$DensePASS, SynPASS$\rightarrow$DensePASS, GTA$\rightarrow$SynPASS, and SynPASS$\rightarrow$ACDC, covering varying degrees of semantic and environmental mismatch. In the real-to-real case, both domains share sunny scenes but include human and vehicle categories that remain private. Synthetic-to-real transfers introduce additional private traffic participants while maintaining shared weather. Synthetic-to-synthetic transfer expands this gap further by adding private meteorological conditions such as cloud, fog, and rain, along with fine-grained static and dynamic categories exclusive to the target. The adverse weather scenario emphasizes extreme condition shifts, where fog and rain remain shared while cloud, sunny, night, and snow are domain-specific. Together, these configurations simultaneously expose category- and weather-level discrepancies, forming a unified and reproducible benchmark for assessing robustness under open-set domain shifts.
\\
\textbf{Input resolution configuration.} For all four open-set benchmark settings, the source and target datasets exhibit diverse image resolutions, reflecting their original collection protocols. To ensure consistency during training, all images are cropped to a unified size of $512{\times}512$. Specifically, CityScapes images are originally $1024{\times}512$ while DensePASS images are $2048{\times}400$, SynPASS images are $2048{\times}1024$, GTA images are $1280{\times}720$, and ACDC images are $960{\times}540$. This unified crop strategy balances the varying aspect ratios and resolutions across datasets, providing the input size for network training while preserving sufficient spatial context for semantic segmentation.
\\

\begin{table*}[h]
\centering
\caption{Comparison of domain adaptation settings with the domain shift.}
\renewcommand{\arraystretch}{1.4}
\setlength{\tabcolsep}{9pt}
\begin{tabular}{>{\raggedright}p{6.8cm}| c |c |c}
\toprule
\rowcolor{gray!10} 
\textbf{Domain Adaptation Setting} & \textbf{Label Shift} & \textbf{Weather Shift} & \textbf{FoV Shift} \\
\midrule
Unsupervised Domain Adaptation~\cite{ganin2015unsupervised}  & \ding{55} & \ding{55} & N/A \\
Panoramic Domain Adaptation~\cite{zheng2024semantics} & \ding{55} & \ding{55} & \ding{51} \\
Open-Set Domain Adaptation~\cite{fang2020open}  & \ding{51} & \ding{55} &N/A \\
Open Compound Domain Adaptation~\cite{liu2020open} & \ding{55} & \ding{51} & N/A \\
\midrule
\rowcolor{blue!15} 
Panoramic Open-Set Domain Adaptation (Ours) & \ding{51} & \ding{51} & \ding{51}  \\
\bottomrule
\end{tabular}
\label{comparing setting}
\end{table*}

\definecolor{BaseGreen}{RGB}{34,139,34}   %
\definecolor{PrivateRed}{RGB}{178,34,34}  %
\begin{table*}[!h]
\centering
\caption{Base and private classes used in the PAN2PAN, SYN2SYN, and SYN2REAL experiments. 
Shared classes (\textcolor{BaseGreen}{green}) are common to both the source and target domains, 
while private classes (\textcolor{PrivateRed}{red}) appear only in either the source or the target domain. 
Weather conditions follow the same convention: \textcolor{BaseGreen}{green} indicates shared conditions across both domains, 
and \textcolor{PrivateRed}{red} indicates conditions specific to a single domain.}
\label{tab:pin2pan_classes}
\adjustbox{width=\linewidth}{
\begin{tabular}{>{\centering\arraybackslash\bfseries}m{3.8cm} | >{\centering\arraybackslash}m{9.5cm} | >{\centering\arraybackslash}m{3cm}}
\toprule
\rowcolor{gray!10} 
\textbf{Setting} & \textbf{Classes (Base / Private)} & \textbf{Weather} \\
\midrule
\makecell{Open-Set  \\ CityScapes $\rightarrow$ DensePASS} & 
\textcolor{BaseGreen}{`road', `sidewalk', `building', `wall', `fence', `traffic light', `vegetation', `terrain', `sky', `car', `bus', `motorcycle', `bicycle'}. \newline
CityScapes/DensePASS: \textcolor{PrivateRed}{`pole', `traffic sign', `person', `rider', `truck', `train'}. &
\textcolor{BaseGreen}{Sunny}$\rightarrow$\textcolor{BaseGreen}{Sunny} \\

\midrule
\makecell{Open-Set \\ SynPASS $\rightarrow$ DensePASS} & 
\textcolor{BaseGreen}{`road', `sidewalk', `building', `wall', `fence', `pole',
    `traffic light', `traffic sign', `vegetation', `terrain',
    `sky', `person', `car'}. \newline
SynPASS: \textcolor{PrivateRed}{`other',  `ground', `bridge', `railtrack', `groundrail', `static', `dynamic', `water'}.
DensePASS: \textcolor{PrivateRed}{`bus', `truck', `train', `motorcycle', `bicycle', `rider'}. &
\textcolor{BaseGreen}{Sunny}$\rightarrow$\textcolor{BaseGreen}{Sunny} \\
\midrule
\makecell{Open-Set \\ GTA $\rightarrow$ SynPASS} & 
\textcolor{BaseGreen}{`road', `sidewalk', `building', `wall', `fence', `pole',
    `traffic light', `traffic sign', `vegetation', `terrain',
    `sky', `person', `car'}. \newline
GTA: \textcolor{PrivateRed}{`rider', `truck', `bus', `train', `motorcycle', `bicycle'}.
SynPASS: \textcolor{PrivateRed}{`other',  `ground', `bridge', `railtrack', `groundrail', `static', `dynamic', `water'}.
&
\textcolor{BaseGreen}{Sunny}$\rightarrow$\textcolor{BaseGreen}{Sunny}, \textcolor{PrivateRed}{Cloud, Fog, Rain, Night} \\
\midrule
\makecell{Open-Set \\ SynPASS $\rightarrow$ ACDC} & 
\textcolor{BaseGreen}{`road', `sidewalk', `building', `wall', `fence', `pole', `traffic light', `traffic sign', `vegetation', `terrain',
    `sky', `person', `car'}. \newline
SynPASS: \textcolor{PrivateRed}{`other',  `ground', `bridge', `railtrack', `groundrail', `static', `dynamic', `water'}.
ACDC: \textcolor{PrivateRed}{`rider', `truck', `bus', `train', `motorcycle', `bicycle'}.
&
\textcolor{BaseGreen}{Fog, Rain, Night, Sunny}, \textcolor{PrivateRed}{Cloud}$\rightarrow$\textcolor{BaseGreen}{Fog, Rain, Night, Sunny}, \textcolor{PrivateRed}{Snow} \\
\bottomrule
\end{tabular}}
\label{dataset_class}
\end{table*}

\begin{table*}[!h]
\centering
\caption{Image resolutions for different datasets involved in the four open-set settings. 
Crop size is unified to $512{\times}512$ for training.}
\label{tab:dataset_image_sizes}
\adjustbox{width=\linewidth}{
\begin{tabular}{>{\centering\arraybackslash\bfseries}m{4.2cm} |
                >{\centering\arraybackslash}m{4.2cm} |
                >{\centering\arraybackslash}m{4.2cm} |
                >{\centering\arraybackslash}m{3cm}}
\toprule
\rowcolor{gray!10} 
\textbf{Setting} & \textbf{Source Image Size } (W$\times$H)& \textbf{Target Image Size} (W$\times$H) & \textbf{Crop Size} (W$\times$H) \\
\midrule

\makecell{Open-Set \\ CityScapes $\rightarrow$ DensePASS} &
CityScapes: 1024$\times$512 &
DensePASS: 2048$\times$400 &
512$\times$512 \\

\midrule
\makecell{Open-Set \\ SynPASS $\rightarrow$ DensePASS} &
SynPASS: 2048$\times$ 1024&
DensePASS: 2048$\times$400 &
512$\times$512 \\

\midrule
\makecell{Open-Set \\ GTA $\rightarrow$ SynPASS} &
GTA: 1280$\times$720 &
SynPASS: 2048$\times$1024 &
512$\times$512 \\

\midrule
\makecell{Open-Set \\ SynPASS $\rightarrow$ ACDC} &
SynPASS: 2048$\times$1024 &
ACDC: 960$\times$540 &
512$\times$512 \\
\bottomrule
\end{tabular}}
\label{dataset_input}
\end{table*}

\begin{table}[t]
\centering
\caption{\textbf{Key Settings of the Proposed EDA-PSeg Framework}}
\label{tab:eda-pseg-config}
\renewcommand{\arraystretch}{1.15} %
\setlength{\tabcolsep}{6pt} %
\resizebox{\linewidth}{!}{
\begin{tabular}{ll}
\toprule
\textbf{Component} & \textbf{Setting} \\
\midrule
Framework & Two-branch open-set UDA: source-supervised + target self-training. \\
Backbone & \textbf{DAFormer} with \textbf{MiT-B5} encoder and Graph decoder. \\
Auxiliary Net & \textbf{MobileSAM} for pseudo-mask refinement. \\
Input Size & $512\times512$ (crop, resize, flip, normalize). \\
Normalization & ImageNet mean=[123.7,116.3,103.5], std=[58.4,57.1,57.4]. \\
Pseudo-labels & Refined by MobileSAM~\cite{zhang2023faster}. \\
Teacher Update & EMA with $\alpha=0.999$. \\
Optimizer & AdamW, LR=$6\times10^{-5}$ (decoder $\times10$). \\
Schedule & Warmup + polynomial decay, 40k iterations. \\
Sampling & DACS~\cite{tranheden2021dacs} with rare-class focus (\texttt{min\_pixels=3000}). \\
Evaluation & H-score \& mIoU. \\
\bottomrule
\end{tabular}
}
\label{Implementation}
\end{table}

\subsection{Implementation Details}
We present the details in \cref{Implementation}, including the pipeline, the auxiliary network, and the training configuration.
\\
\textbf{Pipeline.} Our proposed framework, \textbf{EDA-PSeg}, follows a two-branch pipeline designed for open-set unsupervised domain adaptive semantic segmentation. It consists of a \textit{source-domain supervised branch} and a \textit{target-domain self-training branch}, both of which are processed under a unified data augmentation and preprocessing scheme. 
For the \textit{source domain}, each image and its annotation are loaded and resized before being cropped to a fixed resolution of $512 {\times} 512$. Standard data augmentations, such as random horizontal flipping, are applied to enhance robustness. 
All images are normalized using ImageNet statistics ($mean = [123.675, 116.28, 103.53]$, $std = [58.395, 57.12, 57.375]$) and padded to maintain consistent tensor sizes. 
For the \textit{target domain}, a similar preprocessing strategy is adopted, but with adaptive scale resizing and random cropping to accommodate scale variation and perspective distortion common in real-world scenes. Both source and target domains share identical normalization and padding configurations to ensure consistent feature distribution across domains. 
During training, the model alternately samples mini-batches from both domains following the \textbf{DACS}~\cite{tranheden2021dacs} (Domain Adaptive Cross-domain Self-training) strategy. The source-domain batches provide supervised signals, while the target-domain batches are assigned pseudo-labels that are progressively refined using our auxiliary MobileSAM module (described below). 
At inference time, each test image undergoes deterministic evaluation using a single-scale forward pass with resizing and normalization. Multi-scale and flip testing are disabled for computational consistency. This unified pipeline ensures stable and reproducible adaptation performance while preserving domain-level consistency.
\\
\textbf{Auxiliary Network.} 
Following existing open-set domain adaptation approaches, we adopt \textbf{MobileSAM}~\cite{zhang2023faster} as an auxiliary network to refine pseudo-labels during training. MobileSAM is a lightweight and parameter-efficient variant of the Segment Anything Model (SAM)~\cite{kirillov2023segment}. It is employed solely to distinguish foreground and background regions, improving the reliability of pseudo-labels in the target domain. Specifically, for each target image, MobileSAM generates segmentation masks corresponding to potential foreground classes. 
Class frequencies are then computed from these masks, and the dominant category is selected as the valid pseudo-label class. 
\\
\textbf{Training Configuration.}
We adopt \textbf{DAFormer}~\cite{hoyer2022daformer} as the baseline architecture for UDA semantic segmentation. The model employs a \textbf{MiT-B5~\cite{xie2021segformer}} backbone combined with a {DAFormerHead} decoder variant, modified to predict $14$ semantic categories ($13$ closed-set classes plus one open-set/unknown class). 
Training follows the \textbf{DACS}~\cite{tranheden2021dacs} self-training framework with pseudo-labeling and feature consistency losses. The temporal ensembling coefficient is set to $\alpha=0.999$ to stabilize the teacher-student updates. The feature distance loss is disabled, enabling adaptation to rely primarily on pseudo-label-based consistency rather than ImageNet feature alignment. 
Pseudo-label generation ignores the top $15$ and bottom $120$ pixels of the image to reduce label noise near image borders. 
A rare-class sampling mechanism is incorporated to alleviate class imbalance, using \texttt{min\_pixels $=3000$} and \texttt{class\_temp $=0.01$} to prioritize under-represented categories during source sampling. The model is optimized using the \textbf{AdamW} optimizer with a base learning rate of $6{\times}10^{-5}$, while the learning rate of the decoder head is multiplied by $10$. 
A linear warm-up followed by polynomial decay is applied for stable convergence. Training runs for $40,000$ iterations, and evaluation is performed every $4,000$ iterations. 
Performance is evaluated using both the \textbf{H-score} and \textbf{mIoU}, which together measure segmentation quality and open-set recognition accuracy. The checkpoint achieving the highest \textbf{mIoU} on the validation set is selected as the best. All experiments are performed on a single-GPU setup.

\begin{table}[h]
\centering
\caption{\textbf{Ablation analysis of the EMA module.}}
\resizebox{\linewidth}{!}{
\begin{tabular}{c|l|c|c|c}
\hline
\multicolumn{5}{c}{Cityscapes $\rightarrow$ DensePASS (\textbf{C2D})} \\
\hline
\rowcolor{gray!10}
\textbf{Exp.} & \textbf{Method} & \textbf{Common} & \textbf{Private} & \textbf{H-Score} \\
\hline
\ding{172} & Baseline &52.56 & 8.57 & 14.74 \\
\ding{172} & Self-Att & 55.45 & 10.95 & 18.28  \\
\ding{172} & Self-Att+Margin Projection & 55.64 & 15.06 & 23.71   \\
\ding{173} & EMA  w/o Margin Projection & 55.03&7.56& 13.30 \\
\ding{174} & EMA  w/o Learnable Scale &55.50 &13.07&21.16 \\
\ding{175} & EMA  w/o Learnable Bias &55.59& 7.35&12.99  \\
\ding{176} & EMA  w/o Learnable Magnitude &54.04 &5.20 &9.48 \\
\hline
\rowcolor{blue!15}
\ding{177} & EMA  (Full) &56.12   & 13.00 & 21.11\\
\hline
\end{tabular}
}
\label{eme_abla}
\end{table}

\begin{table}[t]
\centering
\caption{Comparison of the number of parameters (M), FLOPs (G), MACs (G), and test time per image (ms). 
Experiments are conducted on a single RTX 4090 and AMD Ryzen 9 5950X CPU. 
}
\label{111}
\renewcommand{\arraystretch}{1.15}
\setlength{\tabcolsep}{3.5pt}
\begin{tabularx}{\linewidth}{
    >{\raggedright\arraybackslash}X   %
    >{\centering\arraybackslash}X     %
    >{\centering\arraybackslash}X
    >{\centering\arraybackslash}X
    >{\centering\arraybackslash}X
}
\toprule
\rowcolor{gray!10} 
\textbf{Method} & \textbf{\#Params} & \textbf{FLOPs} & \textbf{MACs} & \textbf{Time} \\
\midrule
OSBP~\cite{saito2018open} &85.15 M &59.45 G &29.73 G &26.86 ms \\
UAN~\cite{you2019universal} &85.29 M &59.45 G &29.73 G &26.67 ms \\
UniOT~\cite{jang2022unknown} & 85.22 M &59.45 G&29.73 G&27.03 ms \\
DMLP~\cite{zhang2024behind} &90.15 M&59.45 G&29.73 G&27.12 ms \\
\midrule
MIC~\cite{hoyer2023mic} &85.68 M & 14.86 G & 7.43 G &26.62 ms \\
DAF~\cite{hoyer2022daformer} &85.15 M & 59.45 G &29.73 G & 27.76 ms\\
HRDA~\cite{hoyer2022hrda} & 85.69 M & 14.86 G & 7.43 G & 26.25 ms \\
BUS~\cite{choe2024open} & 85.68 M & 14.86 G & 7.43 G & 26.06 ms \\
\midrule
\rowcolor{blue!15} %
Ours &86.47 M &59.45 G &29.73 G &28.09 ms \\
\bottomrule
\end{tabularx}
\label{Efficiency}
\end{table}

\begin{table}[t]
\centering
\caption{Sensitivity analysis of the self-attention layer in the EMA module. Note that, since the Euler encoding in EMA requires both real and imaginary components, the dimensionality must be even.}
\resizebox{\linewidth}{!}{
\begin{tabular}{c|c|c|c|c} 
\hline
\multicolumn{5}{c}{Cityscapes $\rightarrow$ DensePASS (\textbf{C2D})} \\
\hline
\rowcolor{gray!10}
\textbf{Exp.} & \centering \textbf{Layer} & \textbf{Common} & \textbf{Private} & \textbf{H-Score} \\
\hline
\ding{172} & -    & 52.56 & 8.57 & 14.74 \\
\ding{173} & 2   & 56.12   & 13.00 & 21.11  \\
\ding{174} & 4   &  54.80 & 9.94  &  16.83 \\
\ding{176} & 8  & 54.99  & 6.04 & 10.88 \\
\hline
\end{tabular}
}
\label{Sensitivity_ema}
\end{table}

\begin{table}[t]
\centering
\caption{Sensitivity analysis under different threshold values. 
Results are reported on Cityscapes $\rightarrow$ DensePASS (\textbf{C2D}).}
\resizebox{\linewidth}{!}{
\begin{tabular}{c|c|c|c|c} 
\hline
\multicolumn{5}{c}{Cityscapes $\rightarrow$ DensePASS (\textbf{C2D})} \\
\hline
\rowcolor{gray!10}
\textbf{Exp.} & \centering \textbf{Threshold} & \textbf{Common} & \textbf{Private} & \textbf{H-Score} \\
\hline
\ding{172} & 0.4  & 53.03 & 1.66 & 3.22 \\
\ding{173} & 0.5  & 56.02 & 6.17 & 11.11 \\
\ding{174} & 0.6  & 56.81 & 18.86 & 28.32  \\
\ding{175} & 0.7  & 54.59 &  9.66 &  16.41 \\
\hline
\end{tabular}
}
\label{Sensitivity_threshold}
\end{table}

\begin{figure*}[!t]
    \centering
    \includegraphics[width=1.0\linewidth]{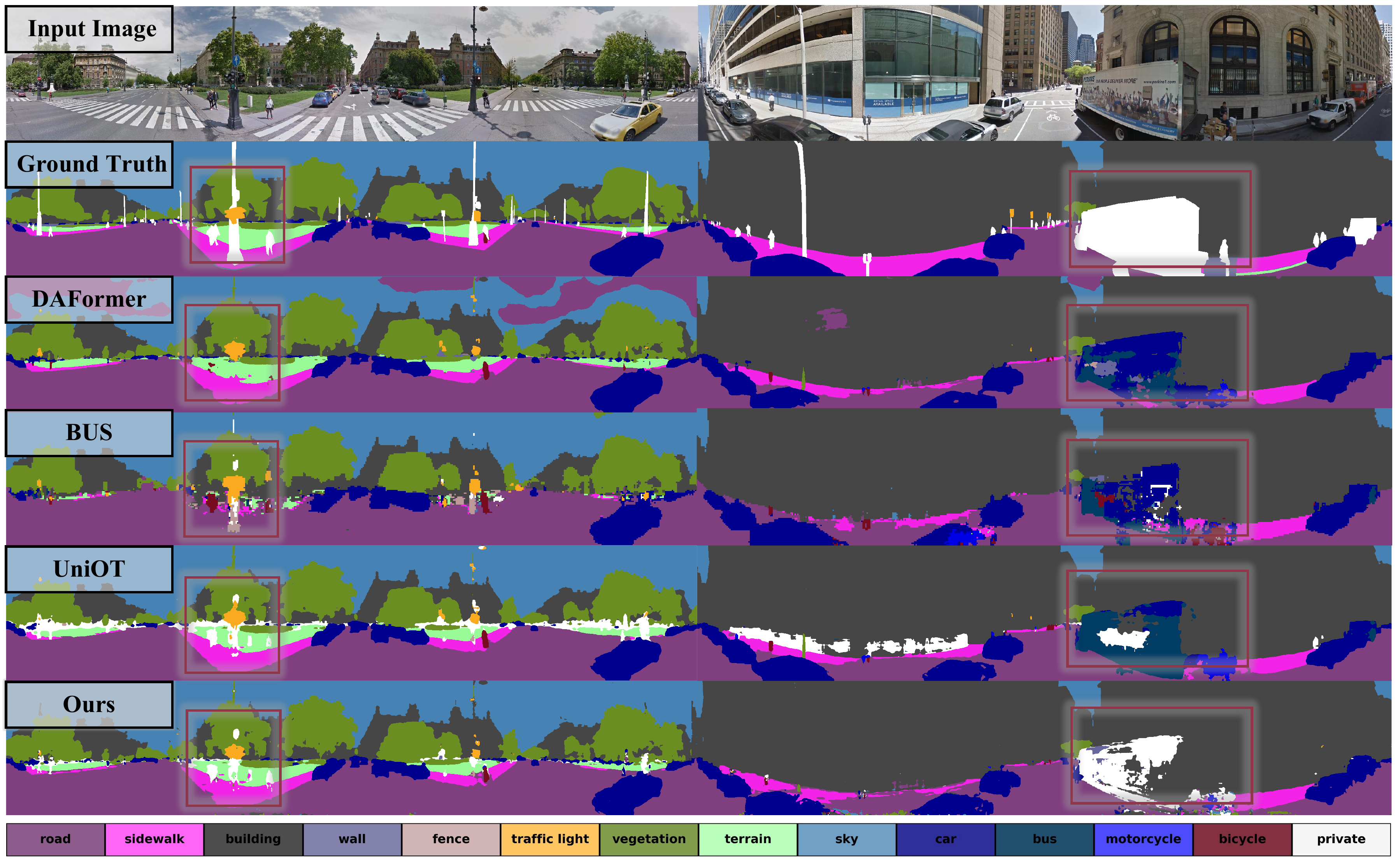}
\caption{Qualitative comparison between our method and existing OSDA and UDA approaches.}
    \label{fig:seg-vis}
\end{figure*}

\begin{figure}[!t]
    \centering
    \includegraphics[width=1.0\linewidth]{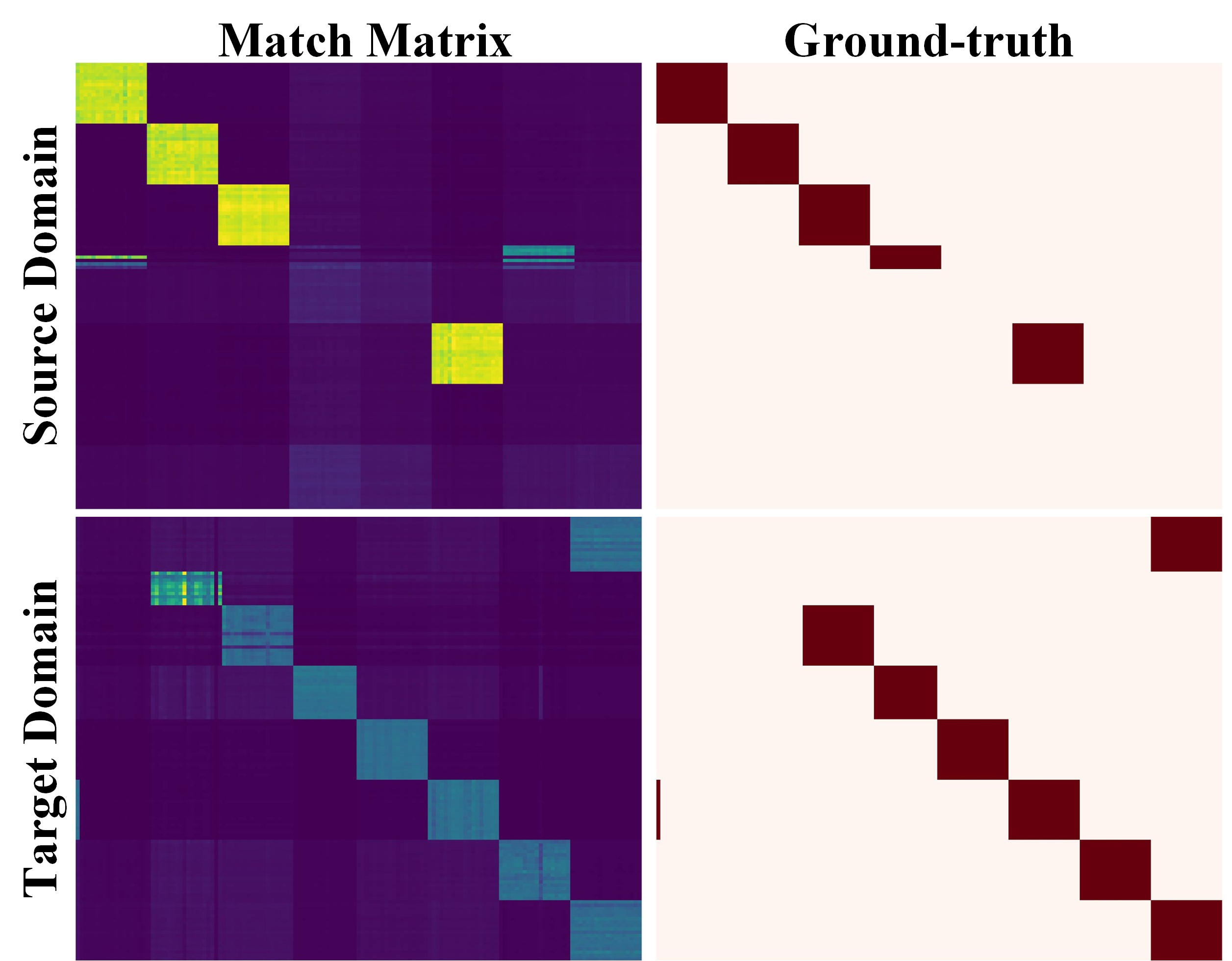}
    \caption{Illustration of the match matrix derived from the graph affinity and the matching ground-truth in the GMA module.}
    \label{graph_match}
\end{figure}

\begin{figure*}
    \centering
    \includegraphics[width=1.0\linewidth]{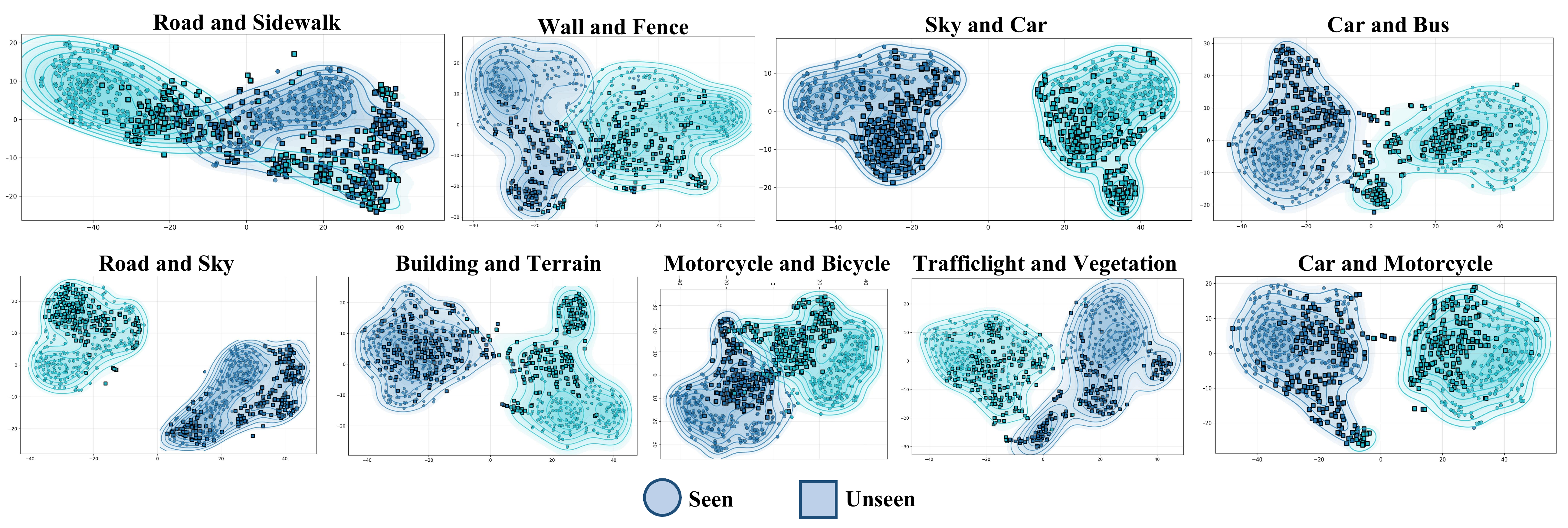}
    \caption{T-SNE visualization of data from seen and unseen viewpoints.}
\label{seen_datasets}
\end{figure*}

\SupSection{Quantitative Comparison}

\subsection{Further Analysis}
As shown in \cref{eme_abla}, we evaluate the effectiveness of the EMA module. The core of our design is the Margin Projection, an enhancement to the self-attention mechanism that consistently improves performance. In the initial comparison (experiment \ding{172}), progressively adding attention-based components consistently improves both Common and Private results, indicating that attention enhances domain-shared representations. Introducing Margin Projection notably increases the Private accuracy and achieves the highest H-Score among non-EMA variants, confirming the benefit of margin-based separation for target-specific learning. In the EMA ablation, removing Margin Projection (\ding{173}) leads to a clear performance drop, suggesting its importance in maintaining target-domain feature separability. The absence of the Learnable Scale (\ding{174}) has a minor effect, while omitting the Learnable Bias (\ding{175}) causes a marked decline, highlighting the need for bias correction to address domain shifts. Finally, removing the Learnable Magnitude (\ding{176}) results in the most severe degradation, demonstrating that the learnable magnitude is crucial for stable feature representation and overall performance.

\subsection{Sensitivity Analysis}
\textbf{Sensitivity of attention layer.} As shown in \cref{Sensitivity_ema}, we conduct the sensitivity analysis to evaluate the effect of self-attention layers in the EMA module. A moderate attention depth notably improves adaptation: two layers yield the highest H-Score, with consistent gains across both common and private classes. This configuration strengthens the angle space coupling induced by the Euler encoding, enabling the model to capture domain-invariant and domain-specific patterns more effectively. However, deeper layers lead to overfitting to source-domain priors, reducing target adaptability and overall balance. These results indicate that the EMA module performs best with limited yet sufficient attention depth, while excessive layers hinder generalization.
\\
\textbf{Sensitivity of threshold.} 
As shown in \cref{Sensitivity_threshold}, we analyze the sensitivity of the threshold used in generating pseudo-labels for private classes during source and target domain mixup training. With the threshold set to $0.4$ (\ding{172}), the model exhibits the lowest mIoU and H-Score for both common and private classes, indicating severe confusion between the two categories. Increasing the threshold to $0.5$ (\ding{173}) leads to a clear improvement in segmentation performance, particularly reflected in a higher H-Score. 
Further raising the threshold to $0.6$ (\ding{174}) enhances the recall of unknown classes while also improving the accuracy on known class mIoU. However, raising the threshold to $0.7$ (\ding{175}) reduces both common and private mIoU, suggesting that too strict a confidence filter can impair pseudo-label generation.

\subsection{Model Efficiency}
As shown in \cref{Efficiency}, we evaluate the efficiency of the parameters (M), FLOPs (G), MACs (G), and test time per image (ms), and compare these metrics with existing methods. OSBP~\cite{saito2018open}, UAN~\cite{you2019universal}, UniOT~\cite{jang2022unknown}, and DMLP~\cite{zhang2024behind} have about $85M$ to $90M$ parameters and $59.45G$ FLOPs, requiring around $27ms$ per image, indicating relatively high complexity. 
In contrast, MIC~\cite{hoyer2023mic}, DAF~\cite{hoyer2022daformer}, HRDA~\cite{hoyer2022hrda}, and BUS~\cite{choe2024open} are more lightweight, with only $14.86$G FLOPs and faster inference. 
Our method achieves a similar computational scale to the above models, with a slightly higher parameter count ($86.47M$) and inference time ($28.09ms$).

\subsection{Visualization Analysis}
\textbf{Qualitative Results.} As illustrated in \cref{fig:seg-vis}, we present a qualitative comparison of open-set UDA segmentation results using the OSDA~\cite{fang2020open} methods BUS~\cite{choe2024open} and UniOT~\cite{jang2022unknown}, as well as the UDA method DAFormer~\cite{hoyer2022daformer}, to evaluate the effectiveness of our proposed approach. 
Compared with existing methods, our approach delivers improved segmentation performance across both known and unknown classes. In particular, it achieves superior open-set performance relative to DAFormer~\cite{hoyer2022daformer}, demonstrates greater robustness to small open-set objects than BUS~\cite{choe2024open}, and substantially mitigates closed-set class and private class foreground objects misidentification errors frequently observed in UniOT~\cite{jang2022unknown}.
\\
\textbf{Graph Match Results.} 
As shown in \cref{graph_match}, we visualize the matching matrix from the learnable affinity and its corresponding ground truth within the proposed GMA module for both the source and target domains. The highlighted regions in the ground truth indicate the nodes that should be matched. The predicted matching matrix closely aligns with the ground truth, demonstrating the effectiveness of the proposed GMA module in open-set graph matching.
\\
\textbf{Seen and Unseen Viewpoints.} 
As shown in \cref{seen_datasets}, we perform visualization experiments on both the seen dataset Cityscapes~\cite{cordts2016cityscapes} and the unseen dataset DensePASS~\cite{ma2021densepass}.
For the stuff classes~\cite{caesar2018coco} (\textit{e.g.}, road, sidewalk, sky) and the thing classes~\cite{caesar2018coco} (\textit{e.g.}, car, bus, motorcycle). For stuff classes and thing classes, categories within the same group tend to share certain similarities, which is reflected in their overlapping or intersecting distributions in the t-SNE visualization. In contrast, when comparing categories across stuff and thing classes, their t-SNE embeddings typically show a significant separation, indicating substantial feature-level differences between the two groups. For objects viewed from both seen and unseen perspectives, domain shift occurs across categories due to variations in the field of view.
\begin{figure*}[!t]
    \centering
    \includegraphics[width=1.0\linewidth]{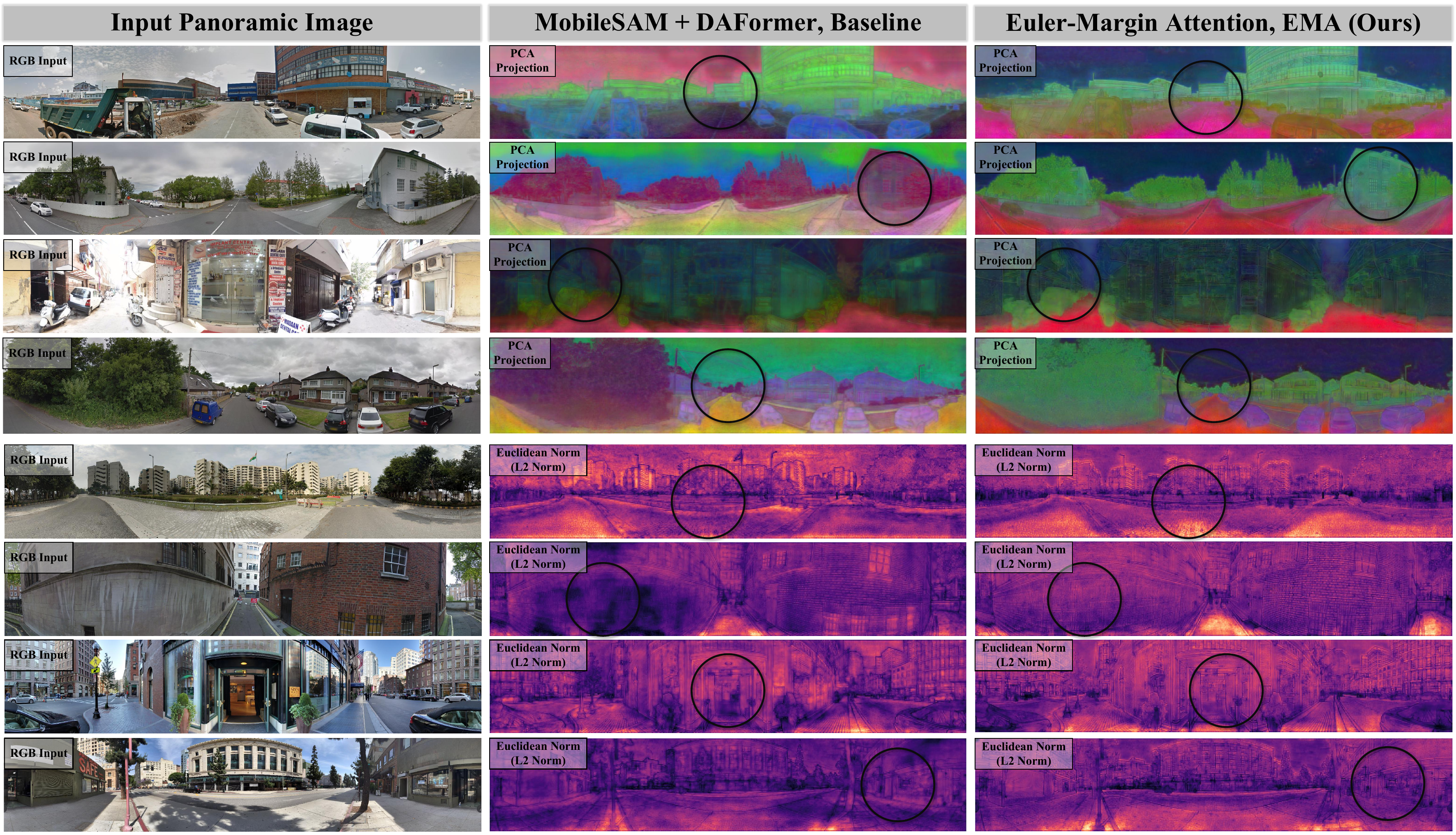}
    \caption{Qualitative results for EMA. We visualize feature representations using Principal Component Analysis (PCA) and the L2 norm.}
\label{ema-vis-all}
\end{figure*}
\\
\textbf{Qualitative results for EMA module.} 
As shown in \cref{ema-vis-all}, we perform qualitative analysis of the EMA module using Principal Component Analysis (PCA) to identify regions with consistent color or brightness, while the L2 norm highlights areas receiving model attention. For the PCA projection, the proposed EMA method better highlights thing classes such as buildings, maintains smoother and less noisy representations in stuff classes like the sky, and achieves higher cross-view consistency in vegetation regions. For the Euclidean (L2) norm visualization, the baseline suffers from limited activation and reduced brightness in geometrically distorted regions, whereas our method effectively overcomes this issue and better highlights objects at long ranges and under wide viewpoints.

\SupSection{Limitations and Outlook}

\begin{figure*}[!t]
    \centering
    \includegraphics[width=1.0\linewidth]{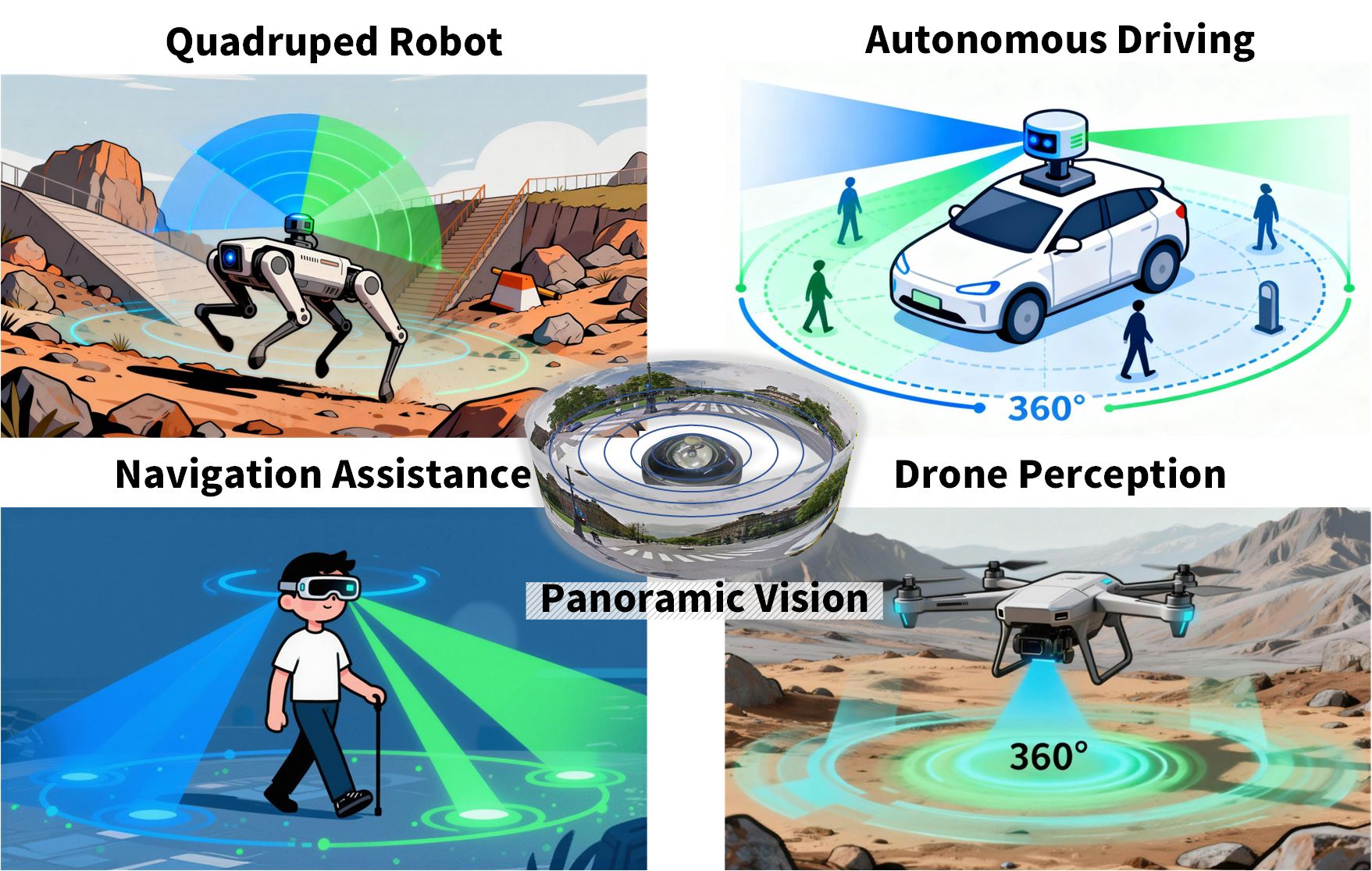}
\caption{Societal implications of panoramic vision technology. 
}
    \label{social}
\end{figure*}

\subsection{Societal Implications}
As illustrated in~\cref{social}, our proposed Extrapolative Domain Adaptive Panoramic Segmentation framework delivers significant societal benefits across four domains: quadruped robotics, autonomous driving, assistive navigation for people with visual impairments, and drone-based perception. 
By enhancing semantic and panoramic perception, our method enables quadruped robots to operate reliably in hazardous environments, supports autonomous vehicles in making safer decisions under diverse conditions, assists visually impaired users with real-time spatial awareness, and strengthens drone perception for environmental monitoring, precision agriculture, and infrastructure inspection.

\subsection{Future Research Directions} 
In the future, we plan to pursue two main research directions. The first focuses on methodological improvements, with the aim of improving the performance, efficiency, and robustness of our current approach. The second direction involves application-oriented extensions, where we intend to adapt and apply our methods to broader or more complex real-world scenarios.
Building upon the first research direction, we note that in pseudo-label training, the private class is currently threshold-based. To advance this, future work suggests improving it to a threshold-free private class mechanism, in line with existing research~\cite{ma2025protogcd,riz2025novel,zhu2024open}.
Despite progress in related areas, test-time adaptation~\cite{liang2025comprehensive,maharana2025batclip} and source-free domain adaptation~\cite{li2024comprehensive,xu2025unraveling} have received limited attention in open-world vision for panoramic images.  
In the future, we will expand the applicability of this method further. Experiments show that the technique maintains stable performance under different fields of view, which confirms its adaptability to various imaging conditions. Subsequent work will explore its applications in panoramic, fisheye, wide-angle, and pinhole images and further extend it to real-world scenes.

\subsection{Limitations and Potential Solutions}
In edge computing scenarios or environments with limited computational resources, our model still exhibits a limit for parameter scalability and inference efficiency. This limitation primarily arises from the adoption of the cross-domain open-set graph matching mechanism and the Euler-Margin attention module. Although these components enhance the generalization in unseen views, they inevitably introduce additional parameter overhead and computational costs during inference. We provide two of the above issues. For model architecture, we consider adopting a lightweight self-attention mechanism integrated with our proposed Euler-interval projection and amplitude \& phase modulation, aiming to preserve representational expressiveness while reducing parameter complexity. To enhance computational efficiency, we plan to conduct the quantization to accelerate inference and reduce memory consumption.

\end{document}